\documentclass[sigconf,authorversion,nonacm]{acmart}
\usepackage[utf8]{inputenc}
\usepackage[T1]{fontenc}
\usepackage{url}
\usepackage{amsfonts}
\usepackage{nicefrac}
\usepackage{microtype}
\usepackage{xcolor}
\usepackage{color}
\usepackage{float}
\usepackage{caption}
\usepackage{subcaption}
\usepackage{wrapfig}
\usepackage{picinpar}
\usepackage{cutwin}
\usepackage{url}
\usepackage{booktabs}
\usepackage{multirow}
\usepackage{epsfig}
\usepackage{graphicx}
\usepackage{amsmath}
\usepackage{bbm}
\usepackage{enumerate}
\usepackage{algorithm}
\usepackage{algpseudocode}
\usepackage{setspace}
\usepackage{bm}
\usepackage{amsthm}
\usepackage{mathtools}
\usepackage{dsfont}
\usepackage{footnote}
\usepackage{lipsum}
\usepackage{makecell}
\usepackage{enumitem}

\newcommand{\indicator}{\mathbbm{1}}
\newcommand\shortsection[1]{\vspace{5pt}{\noindent\bf #1.}}
\newcommand\shortersection[1]{\vspace{3pt}{\noindent\em #1.}}
\newcommand{\RR}{\mathbb R}
\newcommand{\bx}{\bm{x}}
\newcommand{\bw}{\bm{w}}
\newcommand{\g}{\mathrm{g}}
\newcommand{\q}{\mathrm{q}}
\newcommand{\mt}{\mathrm{t}}
\newcommand{\mnt}{\mathrm{nt}}
\newcommand{\targ}{\mathrm{targ}}

\newcommand{\ourAttack}{\text{DynTracker}}
\newcommand{\ourDefense}{\text{DivTrackee}}
\newcommand{\queryImages}{\mathcal{X}_\mathrm{q}}
\newcommand{\galleryImages}{\mathcal{X}_\mathrm{g}}
\newcommand{\galleryIdentities}{\mathcal{Y}_\mathrm{g}}
\newcommand{\galleryDatabase}{\mathcal{D}_\mathrm{g}}
\newcommand{\TrackeeImages}{\mathcal{X}_\mathrm{ee}}
\newcommand{\cosDis}{\mathrm{D}}

\theoremstyle{plain}
\theoremstyle{remark}
\theoremstyle{definition}
\newtheorem{definition}{Definition}

\DeclareMathOperator*{\argmax}{argmax}
\DeclareMathOperator*{\argmin}{argmin}

\begin{document}

\title{DivTrackee versus DynTracker: Promoting Diversity in Anti-Facial Recognition against Dynamic FR Strategy}

\author{Wenshu Fan$^\ast$}
\thanks{$^\ast$These authors contributed equally to this work.}
\thanks{$^\dagger$Corresponding authors.}
\affiliation{\institution{University of Electronic Science and Technology of China}\country{}}
\email{fws@std.uestc.edu.cn}
\author{Minxing Zhang$^\ast$}
\affiliation{\institution{CISPA Helmholtz Center for Information Security}\country{}}
\email{minxing.zhang@cispa.de}
\author{Hongwei Li}
\affiliation{\institution{University of Electronic Science and Technology of China}\country{}}
\email{hongweili@uestc.edu.cn}
\author{Wenbo Jiang$^\dagger$}
\affiliation{\institution{University of Electronic Science and Technology of China}\country{}}
\email{wenbo_jiang@uestc.edu.cn}
\author{Hanxiao Chen}
\affiliation{\institution{University of Electronic Science and Technology of China}\country{}}
\email{chenhanxiao.chx@gmail.com}
\author{Xiangyu Yue}
\affiliation{\institution{The Chinese University of Hong Kong}\country{}}
\email{xyyue@cuhk.edu.hk}
\author{Michael Backes}
\affiliation{\institution{CISPA Helmholtz Center for Information Security}\country{}}
\email{backes@cispa.de}
\author{Xiao Zhang$^\dagger$}
\affiliation{\institution{CISPA Helmholtz Center for Information Security}\country{}}
\email{xiao.zhang@cispa.de}

\renewcommand{\shortauthors}{W. Fan, M. Zhang, H. Li, W. Jiang, H. Chen, X. Yue, M. Backes, X. Zhang}

\begin{abstract}
The widespread adoption of facial recognition (FR) models raises serious concerns about their potential misuse, motivating the development of anti-facial recognition (AFR) to protect user facial privacy.
In this paper, we argue that the static FR strategy, predominantly adopted in prior literature for evaluating AFR efficacy, cannot faithfully characterize the actual capabilities of determined trackers who aim to track a specific target identity.
In particular, we introduce DynTracker, a dynamic FR strategy where the model's gallery database is iteratively updated with newly recognized target identity images.
Surprisingly, such a simple approach renders all the existing AFR protections ineffective.
To mitigate the privacy threats posed by DynTracker, we advocate for explicitly promoting diversity in the AFR-protected images. 
We hypothesize that the lack of diversity is the primary cause of the failure of existing AFR methods.
Specifically, we develop DivTrackee, a novel method for crafting diverse AFR protections that builds upon a text-guided image generation framework and diversity-promoting adversarial losses.
Through comprehensive experiments on various image benchmarks and feature extractors, we demonstrate DynTracker's strength in breaking existing AFR methods and the superiority of DivTrackee in preventing user facial images from being identified by dynamic FR strategies.
We believe our work can act as an important initial step towards developing more effective AFR methods for protecting user facial privacy against determined trackers.
\end{abstract}

\maketitle
\section{Introduction}
\label{section: introduction}

Facial recognition models, where the goal is to recognize the identity of individuals from digital images or videos, have been adopted in various real-world applications such as security checking and attendance management ~\cite{parmar2014face,parkhi2015deep,kortli2020face,adjabi2020past,kaur2020facial}.
Although FR systems greatly enhance the convenience of our daily lives, their potential misuse raises serious concerns.
In particular, after the release of privacy protection regulations such as GDPR~\cite{GDPR} and CCPA~\cite{CCPA}, more people have begun to realize that unauthorized FR models can significantly threaten their facial privacy.
Therefore, there has been growing attention to developing AFR techniques to protect the facial privacy of users who post their images online~\cite{wenger2023sok}.
Existing methods can be mainly divided into two categories, gallery-target AFR~\cite{shan2020fawkes,huang2021unlearnable,cherepanova2021lowkey,chow2024diversity} and query-target AFR~\cite{madry2018towards,dong2018boosting,dong2019evading,yang2021towards,yin2021adv,hu2022protecting,shamshad2023clip2protect,sun2024diffam,le2024styleadv,li2024transferable}, depending on whether perturbations are crafted onto gallery or query images to avoid recognition of the target identity.

Witnessing the lack of a well-defined threat model in the existing literature on AFR, we start by formally defining the problem as a two-party security game between Tracker and Trackee (Section \ref{section: game between tracker and trackee}). We use \textit{Tracker} to denote the malicious user who is determined to track the online images of a specific target identity using FR, while \textit{Trackee} stands for the target identity who deploys AFR protections on their facial images before posting to dodge the tracking. 
Examining relevant literature on anti-facial recognition, we realize all the existing AFR methods, regardless of gallery-target or query-target, adopt a static FR strategy in their evaluation. 

\shortsection{Contribution} 
We argue that such a static assumption largely restricts the Tracker's behavior when using an FR model, which is insufficient to characterize the actual capabilities of determined Trackers (Section \ref{sec: limitation static FR}).
In particular, we propose \textit{DynTracker}, a dynamic FR strategy that iteratively updates the gallery database of Tracker's FR model with newly identified Trackee images (Section \ref{sec: design dynamic FR}). 
DynTracker gradually enriches the gallery database with more clues about Trackee's facial features and the deployed AFR perturbation scheme, enabling it to uncover more of Trackee's facial images.
Based on a preliminary case study, we demonstrate that DynTracker can indeed achieve significantly higher tracking success rates than the static FR strategy, rendering all the existing AFR protection schemes almost completely ineffective (Section \ref{sec: preliminary dyntracker}).

Moreover, we analyze the underlying reasons behind the catastrophic failure of existing AFR methods against our DynTracker, where we hypothesize the main cause is the lack of diversity in the AFR-protected images (Section \ref{subsection: lack of diversity}). 
For generative-based AFR, in particular, a fixed auxiliary image of a different identity from Trackee is typically used to guide the generation process, leading to similar patterns shared among AFR-protected Trackee images. Therefore, once a few protected Trackee images are recognized by the FR model and included in the gallery database, such similarities will significantly decrease the protection success rates of AFR.
To address the privacy risks posed by DynTracker, we develop \textit{DivTrackee}, a novel text-guided generative-based AFR method that builds upon diversity-promoting modules and adversarial losses (Section \ref{subsection: our method}). In particular, DivTrackee explicitly promotes more diverse generations of AFR perturbations by randomly selecting an auxiliary image from a pool of candidate images and penalizing the similarities of newly generated images to previously produced AFR-protected Trackee images stored in a queue.

We conduct extensive experiments across facial image benchmarks and feature extractors to validate DynTracker's strength in identifying Trackee images and DivTrackee's protection efficacy against dynamic FR strategies (Section \ref{section: experiments}). We also vary the Tracker's initial knowledge about the Trackee and consider facial verification models to account for possible variations in Tracker's tracking behaviors. Our results confirm that all the existing AFR methods are highly vulnerable to dynamic FR strategies. For instance, a state-of-the-art generative-based AFR scheme, Clip2Protect~\cite{shamshad2023clip2protect}, achieves only less than $20\%$ protection success rates with respect to Trackee images against our DynTracker.
In sharp contrast, our DivTrackee significantly lowers DynTracker's tracking success rates, consistently achieving more than a $40\%$ increase in protection success while preserving high visual quality of Trackee's AFR-protected images.
Our work highlights the importance of adopting dynamic FR strategies, such as our DynTracker, for more rigorous evaluations of AFR methods and also reveals the effectiveness of promoting more diverse AFR protections as potential countermeasures. We hope our work can serve as an important first step towards developing more reliable AFR protection schemes against determined trackers.

To summarize, our key contributions are as follows:
\begin{itemize}
    \item By characterizing the limitations of static FR strategies, we propose a dynamic FR strategy \textit{DynTracker}, which is easy to implement and more powerful.
    \item To adaptively mitigate the privacy breaches induced by DynTracker, we propose \textit{DivTrackee}, which crafts diverse AFR protections using text-guided image generation and diversity-promoting adversarial losses.
    \item Extensive experiments show that DynTracker effectively breaks existing AFR methods across various facial image benchmarks and feature extractors, and DivTrackee demonstrates superior performance in preventing user facial images from being identified by dynamic FR strategies.
\end{itemize}

\section{Background and Related Work}
\label{section: background and related works}

In this section, we introduce mathematical definitions, necessary preliminaries, and the most relevant literature on facial recognition models and anti-facial recognition technology.

\subsection{Facial Recognition}
\label{subsection: background of facial recognition}
Facial recognition models are designed to identify people by matching their facial characteristics in an unknown image with a database of known faces~\cite{li2020review}, typically consisting of three major components: a pre-trained facial feature extractor, a gallery database of known faces, and a query matching step based on some similarity metric.

\shortsection{Feature Extractor}
Let $\mathcal{X}\subseteq\RR^n$ be the input space of facial images and $\mathcal{Z}\subseteq\RR^d$ be some low-dimensional latent feature space. A feature extractor in a modern FR model is usually a deep neural network $f:\mathcal{X}\rightarrow\mathcal{Z}$ trained to extract distinctive facial features of different identities from their digital images.
Since human faces often contain large variations between identities and even within a single identity, training a high-quality feature extractor from scratch is expensive in terms of both data collection and computational costs.
To encourage collaboration and promote facial recognition applications, various well-established facial feature extractors are publicly available online~\cite{yi2014learning,schroff2015facenet,chen2018mobilefacenets,deng2019arcface,meng2021magface}.
Such open-source efforts provide practitioners and researchers with the feasibility and ease of deploying facial recognition models customized to their needs.

\shortsection{Gallery Database}
In addition to a reliable feature extractor, an FR model requires a collection of gallery images covering a diverse set of identities, denoted as the gallery database $\galleryDatabase=(\galleryImages, \galleryIdentities)$, where $\galleryImages$ is a set of gallery images and $\galleryIdentities$ is their corresponding ground-truth identities. 
Given a target identity that a facial recognition model aims to recognize, an implicit requirement is that the gallery database $\galleryDatabase$ contains at least one annotated facial image of the target identity, a prerequisite for successful facial identification.
Including multiple facial images for the identities within the gallery database can enhance the robustness of facial recognition, particularly when the identity's face images vary widely, which has been adopted in state-of-the-art facial recognition systems.

\shortsection{Query Matching} 
At inference time, a facial recognition model will output an identity prediction for any query image $\bm{x}_{\q}\in\mathcal{X}$ based on the feature extractor $f$, the gallery database $\mathcal{D}_{\g}$, and a query matching step. 
To be more specific, let $\mathcal{Z}_{\g} = \{\bm{z}_{\g} = f(\bm{x}_{\g}): \bm{x}_{\g}\in\galleryImages\}$ be the set of extracted latent features corresponding to the gallery images.
Let $\bm{z}_{\q} = f(\bm{x}_{\q})$ be the extracted latent facial features of $\bm{x}_{\q}$. The query matching step first computes a cosine similarity score between $\bm{z}_{\q}$ and each of the gallery latent features $\bm{z}_{\g}\in\mathcal{Z}_{\g}$ and then outputs the identity $\hat{y}_{\g}$ corresponding to the gallery image with the highest similarity score. 
Let $\mathcal{Y}$ be the output space of identities, then a facial recognition model can be regarded as a function $\mathrm{FR}:\mathcal{X} \rightarrow \mathcal{Y}$ such that: for any query image $\bm{x}_{\q}$,
\begin{align}
\label{equation: facial recognition}
    \mathrm{FR}(\bm{x}_{\q}) = \hat{y}_{\g}, \text{ where} \ (\hat{\bm{x}}_{\g}, \hat{y}_{\g}) = \argmax_{(\bm{x}_{\g}, y_{\g})\in\mathcal{D}_{\g}}  \cos\big(f(\bm{x}_{\q}), f(\bm{x}_{\g})\big).
\end{align}
Note that the FR model, based on the cosine distance from the query image, returns the most similar gallery identity, which is adopted without explicit mention in our paper. 
Using cosine similarity for the query matching step aligns with existing literature~\cite{wang2018cosface,deng2019arcface,meng2021magface}.
Other alternative prediction rules can also be used, depending on the application scenario. For instance, returning a set of top-$k$ most similar identities is also considered in \cite{shamshad2023clip2protect} as the output of the FR model, while Euclidean distance was used to compute the similarity scores in earlier works such as \cite{schroff2015facenet}.
When the gallery database contains a large number of high-dimensional images, dimension-reduction techniques and tree-based structures are often employed in k-nearest neighbor search to improve the computational efficiency of query matching.
In addition, facial verification~\cite{ramanathan2006face,sun2013hybrid,cao2013practical} is closely related to facial recognition, where similar feature extraction and query matching steps are employed.
However, the key difference lies in that a single reference image of the target identity is used to compute the similarity score with a predefined threshold to determine the facial verification result, whereas the gallery database contains multiple images with diverse identities, and no thresholding step is involved in facial recognition.

\subsection{Anti-Facial Recognition}
\label{subsection: background of anti-facial recognition}

Anti-facial recognition is an approach to avoid online tracking of user facial images based on
adversarial examples~\cite{hu2022protecting,wenger2023sok,chow2025personalized}.
An AFR method can be understood as a function $\mathrm{AFR}:\mathcal{X} \rightarrow \mathcal{X}$ that maps any input facial image to a perturbed version of it. Existing AFR techniques can be divided into two categories, \emph{gallery-target} and \emph{query-target}, depending on whether AFR injects small adversarial perturbations into the gallery or query images. 

\shortsection{Gallery-Target}
Gallery-target AFR aims to fool FR models by injecting adversarial perturbations into the gallery images of the target identity. Given a target identity $y_{\mathrm{targ}}$ that AFR aims to protect, the collection of gallery images after perturbation $\tilde\galleryImages$ is defined as:
\begin{equation}
\begin{aligned}
\label{eq:definition gallery protected images}
    \tilde\galleryImages &= \galleryImages^{\mnt} \cup \tilde{\mathcal{X}}_{\mathrm{g}}^{\mt}, \text{ where } \galleryImages^{\mnt} = \{\bm{x}_\mathrm{g}\in\galleryImages \ | \ y_{\g}\neq y_{\mathrm{targ}}\}, 
    \\
    &\quad \text{and } \tilde{\mathcal{X}}_{\mathrm{g}}^{\mt} = \{\mathrm{AFR}(\bm{x}_{\g}) \ | \ \bm{x}_\mathrm{g}\in\galleryImages, \ y_{\g}=y_{\mathrm{targ}}\},
\end{aligned}
\end{equation}
where $\galleryImages$ denotes the set of clean gallery images before perturbation.
Note that only gallery images belonging to $y_{\mathrm{targ}}$ are adversarially perturbed by some gallery-target AFR techniques. 
Due to the injected adversarial features in $\tilde{\mathcal{X}}_{\mathrm{g}}^{\mt}$, an FR model is expected to output an incorrect identity when a clean facial image of the target identity is queried. 

\shortsection{Query-Target}
The other category is known as query-target AFR, which crafts perturbations to the query images of the target. Let $y_{\targ}$ be the target and $y_{\q}$ be the ground-truth identity of $\bm{x}_{\q}$, then the collection of (perturbed) query images $\tilde\queryImages$ is defined as:
\begin{equation}
\begin{aligned}
    \label{eq:definition query protected images}
        \tilde\queryImages &= \queryImages^{\mnt} \cup \tilde{\mathcal{X}}_{\mathrm{q}}^{\mt}, \text{ where } \queryImages^{\mnt} = \{\bm{x}_\mathrm{q}\in\queryImages \ | \ y_{\q}\neq y_{\mathrm{targ}}\} 
        \\
        &\quad \text{and } \tilde{\mathcal{X}}_{\mathrm{q}}^{\mt} = \{\mathrm{AFR}(\bm{x}_{\q}) \ | \ \bm{x}_{\q}\in\queryImages, \ y_{\q}=y_{\mathrm{targ}}\},
\end{aligned}
\end{equation}
where $\tilde{\mathcal{X}}_{\mathrm{q}}^{\mt}$ is the set of perturbed query images with respect to $y_{\mathrm{targ}}$ and some query-target AFR technique, and $\queryImages$ denotes the clean counterpart of $\tilde\queryImages$.
Due to the mismatched features between the unperturbed gallery and perturbed query images, the facial privacy of the target identity can be protected.

\shortsection{Related Work}
In the prior literature on gallery-target AFR, Shan et al. proposed adding small perturbations that pretend to be non-target identities in the latent facial feature space~\cite{shan2020fawkes}, whereas Cherepanova et al. targeted to mislead the commercial facial recognition APIs by creating perturbations based on an ensemble of FR models~\cite{cherepanova2021lowkey}.
Later, Radiya-Dixit et al. argued that there exists an asymmetry between users who posted online facial images and the trainers of FR models, revealing the limitations of gallery-target AFR techniques against future-released FR models~\cite{radiya-dixit2022data}. 
Generally speaking, gallery-target AFR imposes stronger assumptions on both the construction procedure of the FR model's gallery database and when the target identity posts their facial images online with or without AFR protections.
Initial attempts of query-target AFR focused on adversarial perturbations bounded in certain distance metrics such as $\ell_p$-norm~\cite{madry2018towards,dong2018boosting,dong2019evading, oh2017adversarial,yang2021towards,zhong2022opom}, while other works considered patch-based approaches, which typically place an adversarial patch in localized regions~\cite{sharif2019general,komkov2021advhat,xiao2021improving}.
Despite improved privacy protection, these methods suffer undesirable visual artifacts due to the restrictive constraint of the injected perturbations.

Recent works have shifted focus toward generative-based methods to produce more natural adversarial examples.
For instance, Hu et al. proposed AMT-GAN that employs generative adversarial networks for makeup transfer~\cite{hu2022protecting}, Shamshad et al. used text-guided generative models to craft adversarial makeup~\cite{shamshad2023clip2protect}, while Sun et al. developed DiffAM by leveraging the strong generative capability of diffusion models~\cite{sun2024diffam}.
These methods attain state-of-the-art privacy protection while largely preserving the naturalness of original facial images. Due to their superior performance and milder assumptions required, we primarily focus on generative-based AFR in this work.

\section{Security Game Between Tracker and Trackee}
\label{section: game between tracker and trackee}

Recall that our work aims to study AFR technology for facial privacy protection against unauthorized FR models. 
However, existing literature on AFR does not provide a unified definition of attackers and defenders due to their distinct research aims.
To avoid potential confusion, we formally define the problem of anti-facial recognition as a two-party security game between \emph{Tracker} and \emph{Trackee}, which will be used throughout this paper.

\begin{definition}
\label{definition: game between tracker and trackee}
In this security game, Tracker employs an FR model $\mathrm{FR}(\cdot)$ to track the online images Trackee posts.
Before posting, Trackee deploys a query-target AFR protection scheme $\mathrm{AFR}(\cdot)$ on their images to dodge the tracking. Let $y_{\mathrm{ee}}$ denote the Trackee's identity and $\queryImages$ be a collection of clean query images, then the goal of Tracker can be captured by the following optimization problem:
\begin{equation}
    \begin{aligned}
    \label{equation: tracker's goal}
        &\max \sum_{\bm{x}_{\q}\in\queryImages} \indicator \big\{y_{\q} = y_{\mathrm{ee}}, \ \mathrm{FR}\left(\mathrm{AFR}(\bm{x}_{\q})\right) = y_{\mathrm{ee}} \big\}, \\
        &\quad \text{s.t. } \sum_{\bm{x}_{\q}\in\queryImages} \indicator \big\{y_{\q} \neq y_{\mathrm{ee}}, \ \mathrm{FR}(\bm{x}_{\q}) = y_{\mathrm{ee}} \big\} \leq \gamma,
    \end{aligned}
\end{equation}
where $y_{\q}$ stands for the ground-truth identity of query image $\bm{x}_{\q}$, and $\gamma \in\mathbb{N}_{+}$ captures the tolerance threshold of the number of false positives. 
Correspondingly, the goal of Trackee can be defined as:
\begin{align}
\label{equation: trackee's goal}
    \min \sum_{\bm{x}_{\q}\in\queryImages}  \indicator \big\{y_{\q} = y_{\mathrm{ee}}, \ \mathrm{FR} \big( \mathrm{AFR}(\bm{x}_{\q}) \big) = y_{\mathrm{ee}} \big\}.
\end{align}
\end{definition}

\begin{figure}[!t]
    \centering
    \includegraphics[width=0.48\textwidth]{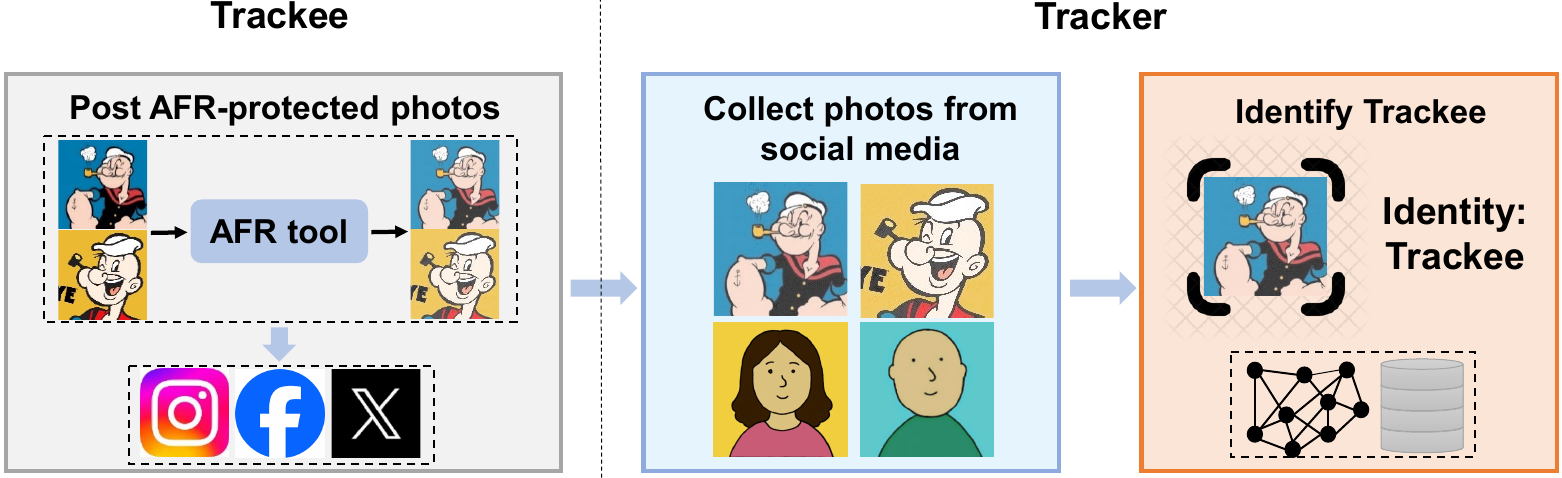}
    \vspace{-0.1in}
    \caption{Illustration of the security game between Tracker and Trackee. Trackee uploads AFR-protected images to social media, while Tracker aims to track Trackee using FR models. To avoid potential ethical issues, we use the fictional cartoon character \emph{Popeye} to represent the victim identity of Trackee.}
    \vspace{-0.05in}
    \label{figure: framework facial recognition}
\end{figure}
Equations \ref{equation: tracker's goal} and Equation \ref{equation: trackee's goal} encode the objectives of Tracker and Trackee, respectively. 
Tracker aims to accurately identify Trackee's images (i.e., maximizing true positives) while avoiding incorrect recognitions of other identities' query images as Trackee (i.e., keeping a low false positive rate). 
In contrast, Trackee aims to lower the identification rate as much as possible by slightly perturbing the posted images using AFR.
Figure \ref{figure: framework facial recognition} illustrates the two-party security game between Tracker and Trackee. 
Note that we consider the query dataset's setting to have multiple different images for each identity, which aligns with the characteristics of the dataset gathered by Tracker in practical scenarios.
To distinguish from the clean query set $\queryImages$, we use $\tilde\queryImages$ (similar in Equation \ref{eq:definition query protected images}) to denote the set of actual query images collected by Tracker. 
In particular, AFR perturbations are assumed to be involved in all the query images of Trackee, while other identities' face images remain unperturbed since Trackees often only have access to their own images.
This assumption largely simplifies our later analyses and is reasonable since, from Trackee's perspective, perturbing all the images before posting is likely to minimize the risk of tracking by Tracker's FR models and maximize protection success rates. 

In this work, we focus on the dodging attack scenario when evaluating the success of AFR, whereas impersonating attacks are often primarily considered in the existing literature on generative-based AFR~\cite{hu2022protecting,shamshad2023clip2protect,sun2024diffam}. The key difference is that dodging aims for an untargeted attack goal through AFR, whereas impersonation is designed to mimic a specific identity different from Trackee, which is a targeted attack. We believe that dodging is a more suitable objective for protecting Trackee's facial privacy than impersonating. As noted in \cite{zhou2024rethinking}, a successful impersonating attack does not necessarily imply high success rates in terms of dodging.

\shortsection{Threat Modeling of Tracker}
We introduce the threat model of the Tracker in more detail. Recall that Tracker's objective is to track Trackee's online posted facial images, where the collection of Trackee's query images is assumed to be all AFR-protected. 
To realize the tracking goal, Tracker is assumed to employ an FR model to recognize Trackee's images from the collected query dataset automatically. 
We assume that Tracker initially has access to a single facial image of the Trackee, which will be included in the gallery database of the adopted facial recognition model.
We believe these assumptions provide a realistic characterization of Tracker's capabilities and align with the expectation that the constraints imposed on Tracker's behavior should be as mild as possible. 

In addition to the above assumptions, we do not impose any constraints on Tracker, aiming to capture the possible variabilities of FR schemes that Trackers may adopt in practice.
In particular, we consider two scenarios for Tracker's initial knowledge, starting with a clean or an AFR-protected image of Trackee, and evaluate the efficacy of AFR with respect to various publicly available FR models and even a black-box facial verification API. 
Existing literature on generative-based AFR only considers the initial knowledge of a clean Trackee gallery image for evaluation. However, since Trackee's online posted AFR-protected image may also be spotted and downloaded by the Tracker, we consider both scenarios of Tracker initially holding a clean or protected Trackee image for comprehensiveness.
In addition, we design a new dynamic FR strategy of Tracker tracking Trackee in Section \ref{section: dynamic strategy of facial recognition}, which is proven to be much stronger than existing static FR schemes and can better capture the actual behavior of determined Trackees and their incentives.

\begin{figure*}[!t]
    \centering
    \includegraphics[width=\textwidth]{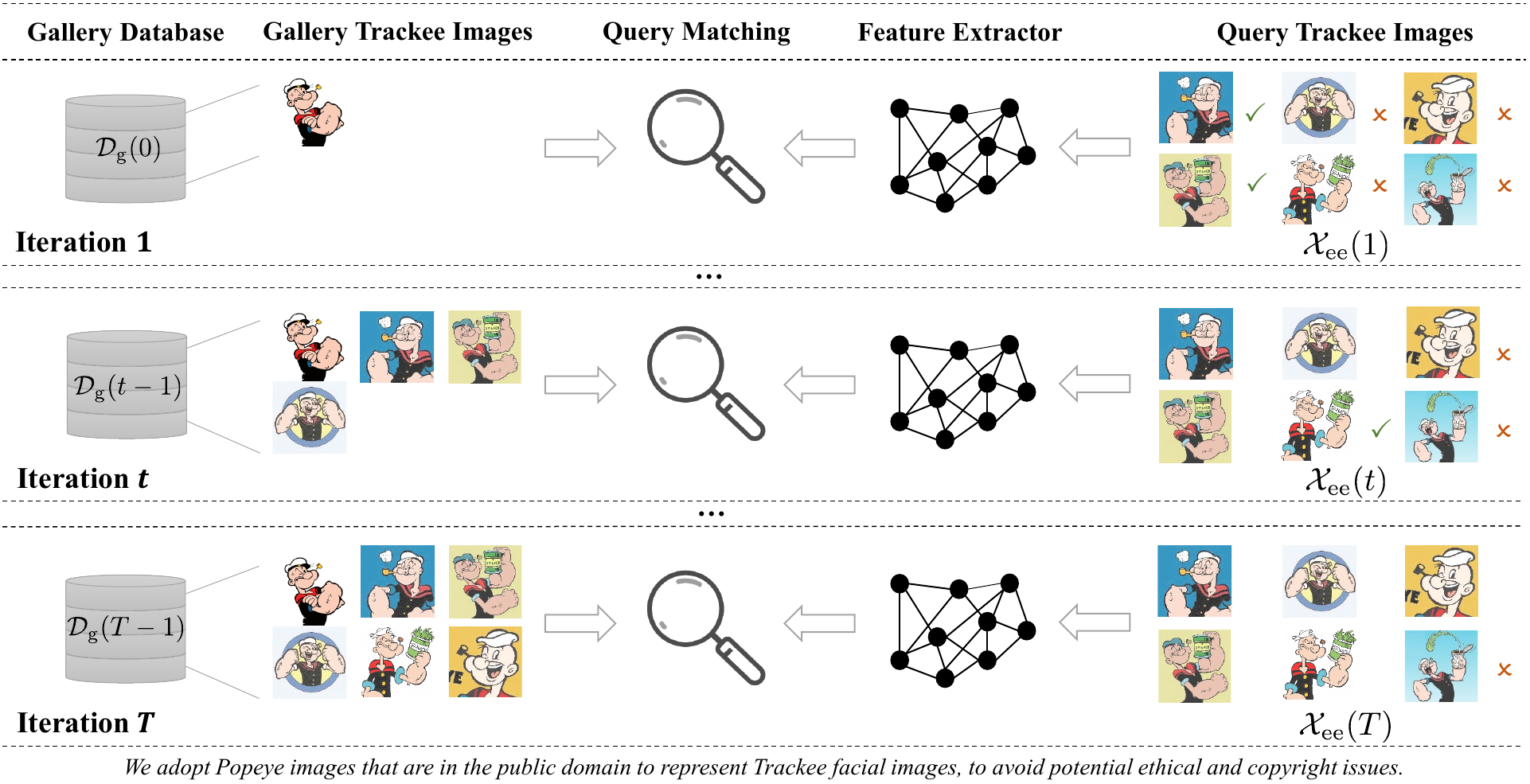}
    \vspace{-0.1in}
    \caption{Illustration of the proposed dynamic facial recognition strategy, DynTracker, for Tracker tracking Trackee.}
    \vspace{-0.05in}
    \label{figure: overview of our dynamic strategy}
\end{figure*}

\shortsection{Expectation for AFR from Trackee} 
Trackee aims to dodge the tracking of unauthorized FR models by crafting AFR perturbations to their images before posting. 
Generally, two main objectives are expected to be attained by a desirable AFR protection scheme: \textit{protection efficacy} and \textit{visual quality}, which will be explained below.

The primary objective in designing AFR is protection efficacy, which characterizes how successfully AFR can avoid identification by FR models. 
Note that Trackee does not have the underlying ground-truth knowledge of the actual FR model employed by Tracker. Thus, it is important to consider the protection efficacy of AFR against various FR models and to promote transferability in the AFR design.
Besides, AFR needs to preserve the visual quality of Trackee's facial images to ensure a satisfactory user experience; otherwise, Trackee may not post the AFR-protected images online if they look unnatural or visually dissimilar to the initial images. We primarily focus on query-target AFR schemes that do not significantly change the visual appearance of Trackee's images, while privacy-preserving tools that remove visually distinctive features of the identity, such as face obfuscation and anonymization~\cite{newton2005preserving,sun2018natural,cao2021personalized}, are considered out of scope.
Since AFR-protected Trackee images are posted before Tracker starts tracking, developing strong AFR schemes is more challenging than crafting adversarial examples to fool ML systems. Trackee has very limited knowledge about the FR model or strategy employed by Tracker. In contrast, traditional adversarial attacks usually assume the victim model can be exploited at least in a black-box manner.
As we will illustrate in Section \ref{section: dynamic strategy of facial recognition}, a simple dynamic FR strategy can render almost all of the existing competitive AFR protection schemes ineffective, confirming the challenges we anticipate for the development of AFR.

\section{Dynamic Strategy of Facial Recognition}
\label{section: dynamic strategy of facial recognition}

In this section, we first explain why the (static) FR strategies adopted in prior AFR literature are insufficient to characterize the capacity of determined Trackers (Section \ref{sec: limitation static FR}). This observation motivates us to develop a dynamic FR strategy, \textit{DynTracker}, which iteratively enriches the gallery database of Tracker's FR model with newly identified Trackee images (Section \ref{sec: design dynamic FR}). In addition, we show preliminary results that DynTracker can significantly lower the protection efficacy of existing AFR methods (Section \ref{sec: preliminary dyntracker}).

\subsection{Motivation}
\label{sec: limitation static FR}

As discussed in Section \ref{section: game between tracker and trackee}, a unique characteristic of the security game between Tracker and Trackee is that AFR protections are deployed on Trackee's images, which are then posted online before Tracker starts the tracking. That said, the constraints imposed on modeling the Tracker's behavior should be as less restrictive as possible. Otherwise, it is likely that we will underestimate the capability of Trackers who are determined to track the Trackee.
Unfortunately, existing gallery-target and query-target AFR methods implicitly assume in their evaluation that the gallery database of the FR models used by Tracker remains fixed. We term such an evaluation setting as the \textit{static FR strategy}. For example, gallery-target AFR injects small perturbations into gallery images of Trackee identity and assumes $\tilde{\mathcal{D}}_{\mathrm{g}}$ is used as the gallery database, while query-target AFR uses a clean facial image dataset with known identities $\galleryDatabase$ as the gallery database. 
Despite different choices, the gallery database will remain unchanged throughout the entire tracking process.

We argue that determined Trackers have both the \textit{capability} and \textit{strong incentives} to update the gallery database dynamically, particularly when the adopted FR model recognizes some of Trackee's query images.
First, given that many well-trained facial recognition models are available online, Trackers can easily download them and make necessary changes to the corresponding gallery database to realize their tracking goal.
In fact, Trackee's initial facial image needs to be inserted into the gallery database to ensure the tracking is specific to the target, which already justifies Tracker's capability of updating the gallery database.
Second, the strong incentives are due to the fact that abundant information about the Trackee might exist in the query images recognized as Trackee. Including these identified Trackee images in the gallery database will likely increase the accuracy of the targeted identity's recognition, thereby matching the goal of determined Trackers.
To support our argument with more concrete evidence, we are going to introduce DynTracker, a dynamic FR strategy featuring multi-round tracking and iterative updates of the gallery database, and provide empirical results showing that Dyntracker is much stronger than the static FR strategy for realizing the tracking goal in the following sections.

\subsection{{\ourAttack}: Dynamic FR Strategy}
\label{sec: design dynamic FR}

So far, we've explained the limitations of existing static FR strategies used in AFR literature and motivated the need to consider dynamic FR strategies for better capturing the capabilities and incentives of determined Trackers. In this section, we introduce the detailed design of our {\ourAttack}, which is illustrated in Figure \ref{figure: overview of our dynamic strategy}. Algorithm \ref{algorithm: DynTracker} in Appendix \ref{appendix: algorithm pseudocode} depicts the pseudocode of our {\ourAttack}.
In particular, {\ourAttack} consists of the following stages of tracking:

\shortsection{Preparation}
Tracker initially holds a single Trackee image, either clean or protected, and includes it in the gallery database of the FR model.
Initialize $\galleryDatabase(0)=\galleryDatabase$ as the gallery database at the start of tracking, and let $\tilde{\queryImages}(0)=\tilde{\queryImages}$ be the set of query images collected by Tracker, consisting of both AFR-protected Trackee images and clean images from other identities. Here, we use $0$ to indicate the $0$-th iteration before any facial recognition happens.

\shortsection{Iterative Update}
As discussed in Section \ref{sec: limitation static FR}, a determined tracker is incentivized to update the gallery database of the FR model with the recognized query images and conduct dynamic facial recognition.
Taking the $t$-th iteration as an example, Tracker first obtains $\TrackeeImages(t)$, the set of query Trackee images predicted as $y_{\mathrm{ee}}$ by the FR model, then includes  $\TrackeeImages(t)$ in the gallery database and perform the $t$-th facial recognition.
To be more specific, the update rule with respect to the $t$-th iteration of DynTracker is defined as: for any $t=1,2,3,\ldots,$
\begin{equation}
\begin{aligned}
\label{equation: update gallery database} 
    \TrackeeImages(t) &= \big\{ \bm{x}_{\q}\in\tilde\queryImages(t-1) \ \big| \ \mathrm{FR} \left( \bm{x}_{\q}; \galleryDatabase(t-1) \right) = y_{\mathrm{ee}} \big\}, \\
    \galleryDatabase(t) &= \galleryDatabase(t-1)\cup \left\{(\bx, y_{\mathrm{ee}}) \ \big| \ \bx \in \TrackeeImages(t) \right\}, \\
    \tilde\queryImages(t) &= \tilde\queryImages(t-1) \setminus \TrackeeImages(t),
\end{aligned}
\end{equation}
where identified images from the previous iteration are dynamically removed from the query dataset to avoid repetition, and we explicitly write out the dependence of $\mathrm{FR}(\cdot)$ on the corresponding gallery database for clarity.
Since $\galleryDatabase(t)$ is updated dynamically, the set of Trackee query images $\TrackeeImages(t)$ returned by the FR model will be different for different iterations. Technically speaking, there might exist other identities' images incorrectly recognized as Trackee in the returned set $\TrackeeImages(t)$. However, the false positive rate turns out to be very low if the FR model is well-trained, which has been verified in our experiments (see Section \ref{subsection: false positive rate} for more discussions on the negligible impact of false positive rates). This iterative update stage gradually inserts more Trackee images into the gallery database to improve the target identity's recognition.

\shortsection{Completion}
As long as new query images can be recognized as Trackee's identity $y_{\mathrm{ee}}$, Tracker will continue the iterative update stage, aiming to uncover as many Trackee images as possible.
Finally, Tracker will complete the whole tracking process when no more Trackee query images are returned from the previous iteration since no benefits can be gained from an extra FR step. For ease of presentation, we use $T$ to denote the total number of iterations.

Although the gallery dataset is dynamically updated during tracking, the query matching scheme remains the same, which still outputs the identity of the gallery image most similar to the query image.  
As more Trackee images are recognized and included in the gallery database, the likelihood of the remaining Trackee query images identified by the updated FR model is expected to increase.
It is worth noting that the first iteration of the dynamic update phase corresponds to the static FR strategy used in existing query-target AFR methods; thus, our DynTracker is more powerful by design in identifying Trackee's query facial images.

\subsection{DynTracker Defeats Existing AFR}
\label{sec: preliminary dyntracker}

Moreover, we conduct a preliminary study to evaluate the efficacy of the proposed dynamic strategy {\emph\ourAttack} with comparisons to the static FR strategy. 
In particular, we randomly select an identity from FaceScrub~\cite{ng2014data} as the Trackee. Tracker employs ResNet-101~\cite{he2016deep} as the feature extractor, which is trained on MS-Celeb-1M~\cite{guo2016ms} via MagFace loss~\cite{meng2021magface}, and adopts FaceScrub as the gallery database with a single clean or protected image of Trackee. 
Table \ref{table: case study of dynamic attack strategy} summarizes the tracking success rates against various AFR methods, including gallery-target schemes~\cite{shan2020fawkes,cherepanova2021lowkey}, query-target approaches based on adversarial noises~\cite{madry2018towards,dong2018boosting,dong2019evading,yang2021towards} and generative models~\cite{yin2021adv,hu2022protecting,shamshad2023clip2protect}. 
We strictly follow the iterative update rule specified in Section \ref{sec: design dynamic FR} to carry out {\ourAttack} against query-target AFR schemes. In comparison, we follow a similar dynamic strategy to evaluate gallery-target AFR but iteratively update the protected gallery database $(\tilde{\mathcal{X}}_{\g}, \mathcal{Y}_{\g})$ with identified clean query images recognized as Trackee. This aligns with the evaluation setup of existing works on gallery-target AFR \cite{shan2020fawkes,cherepanova2021lowkey} (see Section \ref{subsection: background of anti-facial recognition} for details).

\begin{table}[!t]
\centering
\caption{Comparisons of tracking success rates $(\%)$  between static and dynamic FR strategies against various AFR methods. Here, \textit{Clean} and \textit{Protected} indicate Tracker starting with a single clean and protected Trackee image, respectively.}
\vspace{-0.05in}
\resizebox{0.48\textwidth}{!}{
\small
\centering
\begin{tabular}{l | c c | c c}
    \toprule
    \multirow{2.4}{*}{\textbf{AFR Method}} & \multicolumn{2}{c|}{\textbf{Clean}} & \multicolumn{2}{c}{\textbf{Protected}} \\
    \cmidrule{2-5}
    & \textbf{Static} & \textbf{Dynamic} & \textbf{Static} & \textbf{Dynamic} \\
    \midrule
    No Protection & $96.15$ & $99.62$ & N/A & N/A \\
    \midrule
    Fawkess~\cite{shan2020fawkes} & N/A & N/A & $38.75$ & $98.87$ \\
    Lowkey~\cite{cherepanova2021lowkey} & N/A & N/A & $9.37$ & $98.69$ \\
    \midrule
    PGD~\cite{madry2018towards} & $28.47$ & $98.76$ & $48.32$ & $97.66$ \\
    MI-FGSM~\cite{dong2018boosting} & $31.93$ & $98.94$ & $47.36$ & $97.53$ \\
    TI-DIM~\cite{dong2019evading} & $25.57$ & $97.80$ & $45.87$ & $96.68$ \\
    TIP-IM~\cite{yang2021towards} & $13.82$ & $97.10$ & $40.56$ & $96.54$ \\
    \midrule
    Adv-Makeup~\cite{yin2021adv} & $43.49$ & $98.02$ & $62.34$ & $97.12$ \\
    AMT-GAN~\cite{hu2022protecting} & $44.25$ & $99.87$ & $83.50$ & $98.85$ \\ 
    Clip2Protect~\cite{shamshad2023clip2protect} & $33.83$ & $98.04$ & $80.39$ & $98.04$ \\
    DiffAM~\cite{sun2024diffam} &$26.74$ &$98.10$ &$82.36$ &$98.42$ \\
    \bottomrule
\end{tabular}
}
\vspace{-0.05in}
\label{table: case study of dynamic attack strategy}
\end{table}

Table \ref{table: case study of dynamic attack strategy} shows that {\ourAttack} significantly outperforms the static FR strategy in identifying Trackee images from the query dataset, regardless of the initial Tracker's knowledge being a single clean or protected Trackee image.
For instance, starting with a single clean image, {\ourAttack} can increase the tracking success rate from $33.83\%$ to $98.04\%$ against Clip2Protect, almost rendering the deployed AFR protection useless.
Even against the static FR strategy, if Tracker initially holds a protected Trackee image, state-of-the-art generative-based AFR methods exhibit significantly decreased protection efficacy;
not to say against our proposed dynamic FR strategy, all the existing AFR methods appear to be almost useless.
Our preliminary results confirm that the static FR strategy is insufficient to capture the capability of determined Trackers, where the stronger DynTracker should be considered when evaluating the protection effectiveness of AFR methods (see Table \ref{table: main results} and Table \ref{table: main results full} for more comprehensive results with similar trends).

\section{Promoting Diversity to Combat DynTracker}
\label{section: promoting diversity protects facial privacy}

As demonstrated in Section \ref{sec: preliminary dyntracker}, existing AFR methods are reasonably effective against the static FR strategy but fail against our {\ourAttack}.
In this section, we hypothesize that a fundamental reason behind such a failure is the lack of diversity in the design of AFR protections (Section \ref{subsection: lack of diversity}), which further motivates us to develop \textit{DivTrackee}, a novel framework that explicitly promotes diversity in generating AFR protections for Trackee images (Section \ref{subsection: our method}).
Algorithm \ref{algorithm: DivTrackee} in Appendix \ref{appendix: algorithm pseudocode} presents the pseudocode of DivTrackee.

\subsection{Lack of Diversity in Existing AFR}
\label{subsection: lack of diversity}
Before developing countermeasures, it is important to pinpoint the cause of failure for existing AFR against DynTracker.
As explained in Section \ref{sec: design dynamic FR}, the main difference between the static FR strategy and DynTracker lies in the set of gallery images corresponding to Trackee's identity.
As DynTracker proceeds with more iterative updates, more query images recognized as Trackee will be included in the gallery database.
Compared with $\galleryDatabase(0)$, the gallery database updated after the $t$-th iteration of DynTracker $\galleryDatabase(t)$ is expected to contain multiple AFR-protected images of Trackee, thus covering more variations of Trackee's facial features and may provide crucial information about the deployed AFR protection scheme, both of which increase the success rates of Tracker tracking Trackee.

We hypothesize that adversarial perturbations produced by existing AFR share a similar pattern across different facial images, causing the failure against dynamic FR strategies.
Revisiting Table \ref{table: case study of dynamic attack strategy}, we can observe that protection success rates across different query-target AFR methods against the static FR strategy are consistently lower when the Tracker starts with a protected Trackee image, compared with the other clean settings. 
Such a uniformly decreased performance supports our hypothesis that adversarial perturbations generated by a specific AFR scheme are likely to share similar characteristics, which DynTracker can further exploit. 
For generative-based AFR~\cite{yin2021adv,hu2022protecting,shamshad2023clip2protect,sun2024diffam}, an auxiliary facial image that belongs to a different person, denoted as $\bm{x}_{\mathrm{aux}}$ for simplicity, is usually employed to guide the targeted image generation process.
However, the same auxiliary image is shared across different query images when generating AFR perturbations, potentially leading to low protection effectiveness against our DynTracker.
For traditional adversarial noise-based AFR~\cite{madry2018towards,dong2018boosting,dong2019evading,yang2021towards}, although no auxiliary image $\bm{x}_{\mathrm{aux}}$ is utilized, 
they rely on untargeted adversarial examples, which may inherently be prone to misclassification into a particular identity class close to Trackee in the latent space.
We believe the failure of existing AFR methods in privacy protection against DynTracker can be mainly attributed to the lack of explicit diversity promotion in protected image generations.
As shown in Table \ref{table: case study of dynamic attack strategy}, existing AFR methods fail catastrophically against our DynTracker, which provides strong empirical support to our hypothesis.

\subsection{{\ourDefense}: Diversity-Promoting AFR}
\label{subsection: our method}

Built upon previous analyses, we explore whether promoting diversity in AFR can result in more effective protection.
In the following discussions, we will focus on generative-based AFR since it can preserve a better visual quality of facial images than adversarial noise-based AFR. Besides, unlike gallery-target AFR that assumes Trackers will construct their FR model's gallery database in a certain period of time~\cite{radiya-dixit2022data}, generative-based AFR does not impose strong assumptions on Tracker's behavior to be effective. 

\shortsection{Initial Attempt}
Our first attempt is to promote diversity by assigning different Trackee images with distinct auxiliary images in existing generative-based AFR.
Specifically, let $\mathcal{X}_{\mathrm{aux}}$ be the auxiliary image dataset with distinct identities.
For any clean Trackee image $\bm{x}_{\mathrm{ee}}$, one can randomly select an auxiliary image $\bm{x}_{\mathrm{aux}}\in\mathcal{X}_{\mathrm{aux}}$, then apply some generative-based AFR method to craft the protection on $\bm{x}_{\mathrm{ee}}$. 
Since randomly selected auxiliary images with different identities are used to guide the image generation process, the diversity of Trackee's protected images is expected to increase.

Under the same setting as our preliminary experiments in Section \ref{sec: preliminary dyntracker}, we evaluate whether diversifying the auxiliary images can help improve AFR performance.
We select $2000$ images of distinct identities from LFW~\cite{huang2008labeled} to construct the auxiliary dataset $\mathcal{X}_{\mathrm{aux}}$.
Table~\ref{table: initial attempt of diversity} shows the evaluation results for a few makeup-based AFR methods with diversified auxiliary images.
Compared with fixing the auxiliary image, diversifying auxiliary images increases the protection success
of generative-based AFR against {\ourAttack}. 
In particular, tracking success rates of {\ourAttack} against Clip2Protect are lowered by $15.39\%$ and $12.23\%$, respectively, when Tracker initially holds a single clean and protected Trackee image. For DiffAM, diversifying auxiliary images improves performance by nearly $20\%$ against DynTracker.
Nevertheless, the limited overall improvement in protection effectiveness is unsatisfactory from Trackee's perspective.
Therefore, we take a further step to explore the concept of diversity and propose \textit{DivTrackee}, a novel generative-based AFR method with diversity-promoting modules and losses to enhance the protection efficacy of Trackee images against DynTracker.

\begin{table}[!t]
\centering
\caption{Comparisons of {\ourAttack}'s tracking success rates $(\%)$ between \textit{fixed} and \textit{diverse} auxiliary images for generative-based AFR. Here, \textit{Clean} and \textit{Protected} mean Tracker starts with a single clean and protected Trackee image, respectively.}
\vspace{-0.05in}
\resizebox{0.48\textwidth}{!}{
\small
\centering
\begin{tabular}{l | c c | c c}
    \toprule
    \multirow{2.4}{*}{\textbf{AFR Method}} & \multicolumn{2}{c|}{\textbf{Clean}} & \multicolumn{2}{c}{\textbf{Protected}} \\
     \cmidrule{2-5}
    & \textbf{Fixed} & \textbf{Diverse} & \textbf{Fixed} & \textbf{Diverse} \\
    \midrule
    Adv-Makeup~\cite{yin2021adv} & $98.02$ & $96.23$ & $97.12$ & $95.82$ \\
    AMT-GAN~\cite{hu2022protecting} & $99.87$ & $97.69$ & $98.85$ & $92.54$ \\
    Clip2Protect~\cite{shamshad2023clip2protect} & $98.04$ & $82.65$ & $98.04$ & $85.81$ \\
    DiffAM~\cite{sun2024diffam} &$98.10$ &$79.68$ &$98.42$ &$82.90$ \\
    \bottomrule
\end{tabular}
}
\vspace{-0.1in}
\label{table: initial attempt of diversity}
\end{table}

\shortsection{Text-Guided AFR Generation}
We adopt StyleGAN~\cite{KarrasLA19stylgan}, which has been employed for various text-guided image generation and editing tasks~\cite{richardson2021encoding,tov2021designing,yin2022styleheat,sauer2022stylegan,shamshad2023clip2protect}, as the generative backbone of DivTrackee.
Let $G_{\theta}:\mathcal{W}\rightarrow\mathcal{X}$ be a generative model that maps a latent vector $\bm{w}\in\mathcal{W}$ to a generated facial image $G_{\theta}(\bm{w})$.
Given a Trackee's clean image $\bm{x}_{\mathrm{ee}}$ and a target text prompt $p_{\mathrm{targ}}$ (e.g., ``natural makeup''), the goal is to search for a latent code $\hat{\bm{w}}\in\mathcal{W}$ such that the generated image $G_{\theta}(\hat{\bm{w}})$ follows the specifications of the text prompt $p_{\mathrm{targ}}$ and satisfies the expectations for AFR from Trackee discussed in Section \ref{section: game between tracker and trackee}. Using text-guided generative models and makeup prompts allows ease and flexibility in creating natural AFR perturbations on Trackee's facial images.

To obtain a desirable $\hat{\bm{w}}$, we start with finding a good initialization $\bm{w}_{\mathrm{init}}$ such that $G_{\theta}(\bm{w}_{\mathrm{init}})$ is close to $\bm{x}_{\mathrm{ee}}$ to ensure visual similarity.
Leveraging the inversion-based technique for StyleGAN image manipulation~\cite{tov2021designing}, we use a pre-trained e4e encoder $F_\phi:\mathcal{X}\rightarrow\mathcal{W}$ to infer the inverse of the clean Trackee image as the initialization $\bw_{\mathrm{init}} = F_\phi(\bm{x}_{\mathrm{ee}})$.
Inspired by \cite{roich2022pivotal,shamshad2023clip2protect}, we fine-tune the generator $G_{\theta}$ to improve the quality of the reconstructed image $G_{\theta}\big(F_\phi(\bm{x}_{\mathrm{ee}})\big)$ without sacrificing the editing capabilities of the generative model (see Appendix \ref{appendix: generator fine-tuning} for implementation details of such a generator fine-tuning step).
After acquiring the fine-tuned generator, denoted as $G_{\theta^*}$, we perform gradient descent on a carefully designed loss function $\mathcal{L}_\mathrm{tot}$ to optimize $\bm{w}_{\mathrm{init}}$. To be more specific, the gradient update rule is defined as: for any $s=0,1,2,\ldots,S-1$,
\begin{align}
\label{equation: gradient descent}
    \bm{w}_{s+1} = \bm{w}_{s} - \lambda\nabla_{\bm{w}}\mathcal{L}_\mathrm{tot}(\bm{w}_{s}).
\end{align}
Here $\bm{w}_{0}$ is initialized as $\bm{w}_\mathrm{init}$, $S$ is the total number of gradient updates, and $\lambda$ denotes the step size. In particular, $\mathcal{L}_\mathrm{tot}$ represents the total loss function that encodes all Trackee's expectations for text-guided AFR generation, which will be detailed next.

\shortsection{Diversity-Promoting Adversarial Loss}
Now that we've determined the generative framework, the remaining task is to design the total loss $\mathcal{L}_{\mathrm{tot}}$ such that the final output from the gradient-based optimization $G_{\theta^*}(\bm{w}_{S})$ achieves high protection success rates against DynTracker and maintains the visual quality with reference to the Trackee's original image $\bm{x}_{\mathrm{ee}}$ and the given text prompt $p$. To achieve high protection efficacy against DynTracker, we design the following adversarial loss that explicitly promotes diversity:
\begin{align}
\label{equation: protection efficacy loss}
    &\mathcal{L}_{\mathrm{eff}}(\bm{w}) = \mathcal{L}_{\mathrm{adv}}(\bm{w}) + \alpha_1 \cdot \mathcal{L}_{\mathrm{guide}}(\bm{w}) + \alpha_2 \cdot \mathcal{L}_{\mathrm{div}}(\bm{w}),
\end{align}
where $\alpha_1 \geq 0$ and $\alpha_2 \geq 0$ are hyperparameters that control the weights of each loss term. In particular, $\mathcal{L}_{\mathrm{eff}}$ consists of three terms: an adversarial loss $\mathcal{L}_{\mathrm{adv}}$, a diverse guidance loss $\mathcal{L}_{\mathrm{guide}}$ and a diversity-promoting loss $\mathcal{L}_{\mathrm{div}}$, which are detailed below:

\shortersection{Adversarial Loss}
The first loss term ${\mathcal{L}_\mathrm{adv}}$ in Equation \ref{equation: protection efficacy loss} contributes to the primary goal of dodging: avoiding $G_{\theta^*}(\bm{w})$ being recognized as Trackee's identity.
Specifically, let $f$ be a facial feature extractor, then the adversarial loss ${\mathcal{L}_\mathrm{adv}}$ is defined as:
\begin{align}
\label{equation: adversarial loss}
    \mathcal{L}_{\mathrm{adv}}(\bm{w}) = -\cosDis_f \big( G_{\theta^*}(\bm{w}), \bm{x}_{\mathrm{ee}} \big),
\end{align}
where $\cosDis_f: \mathcal{X} \times \mathcal{X} \rightarrow [0, 2]$ captures the cosine dissimilarity between the features of two input images extracted by the feature extractor $f$, i.e., $\cosDis_f\big(\bm{x}_1,\bm{x}_2\big)= 1 - \cos\big(f(\bm{x}_1),f(\bm{x}_2)\big)$.
Note that the feature extractor used in $\mathcal{L}_{\mathrm{adv}}$ does not need to be the same as that of {\ourAttack}.
As we optimize $\bm{w}$ to decrease $\mathcal{L}_{\mathrm{adv}}$ through gradient descent, $G_{\theta^*}(\bm{w})$ is expected to be less similar to the original Trackee image $\bx_{\mathrm{ee}}$ in the latent facial feature space.

\shortersection{Diverse Guidance Loss}
The second term $\mathcal{L}_{\mathrm{guide}}$ in Equation \ref{equation: protection efficacy loss}  denotes the diverse guidance loss, which encourages the generated image $G_{\theta^*}(\bm{w})$ to get closer to a randomly selected auxiliary image $\bm{x}_{\mathrm{aux}}$ up to some margin-based threshold $\delta\geq 0$.
Specifically, the diverse guidance loss $\mathcal{L}_{\mathrm{guide}}$ is defined as:
\begin{align}
\label{equation: region loss}
    \mathcal{L}_{\mathrm{guide}}(\bm{w}) = \max \big( 0,\cosDis_f \left( G_{\theta^*}(\bm{w}),\bx_{\mathrm{aux}} \right) -\delta \big).
\end{align}
Similar to our initial attempt, different Trackee image $\bm{x}_{\mathrm{ee}}$ will be assigned with random auxiliary image $\bm{x}_{\mathrm{aux}}$ selected from a dataset $\mathcal{X}_{\mathrm{aux}}$ to guide the generation. We additionally employ a margin-based loss with a threshold $\delta$ to allow less restrictive directions to search for the desirable latent code.

\shortersection{Diversity-Promoting Loss}
The last loss term ${\mathcal{L}_{\mathrm{div}}}$ in Equation \ref{equation: protection efficacy loss} explicitly promotes diversity within the generated protected images of Trackee.
Specifically, a queue $\mathcal{Q}$ with maximum length $m>0$ is maintained using the \textit{first-in, first-out} (FIFO) principle. 
For any newly arrived Trackee's query image $\bm{x}_{\mathrm{ee}}$, $\mathcal{L}_{\mathrm{div}}$ is defined as the averaged cosine dissimilarity to previously stored images in $\mathcal{Q}$:
\begin{align}
\label{equation: diversity loss}
    \mathcal{L}_{\mathrm{div}}(\bm{w}) = - \frac{1}{m} \sum_{\bm{x}\in\mathcal{Q}} \cosDis_f\big(G_{\theta^*}(\bm{w}), \bm{x}\big).
\end{align}
Note that $\mathcal{Q}$ is sequentially expanded as we generate Trackee's protected images. 
When the first Trackee image arrives, $\mathcal{Q}$ is empty, so the initial $\mathcal{L}_{\mathrm{div}} = 0$. After the first protected image is generated and stored in $\mathcal{Q}$, the generation of the new image needs to account for previous ones. When $|\mathcal{Q}|$ reaches the maximum allowable length $m$, the queue $\mathcal{Q}$ will be updated in a FIFO manner. In our experiments, we set $m=10$ to achieve the diversity-promoting goal without incurring high computational overhead, where further increasing the value of $m$ does not improve the performance much.
Minimizing $\mathcal{L}_{\mathrm{div}}$ encourages the newly generated image $G_{\theta^*}(\bm{w})$ to be dissimilar to the previously-stored protected Trackee images, thus promoting diversity in AFR and enhancing the protection against DynTracker.
As will be illustrated in Section \ref{subsection: effectiveness of loss terms}, the proposed diversity-promoting loss terms, $\mathcal{L}_{\mathrm{guide}}$ and $\mathcal{L}_{\mathrm{div}}$, are essential for DivTrackee to achieve high protection success rates. 

\shortsection{Loss Design for Visual Quality}
In addition to designing diversity-promoting adversarial loss necessary to achieve effective protection against DynTracker, we also need to ensure the visual quality of generated images. To achieve this goal, we adopt the following loss inspired by prior work~\cite{gal2022stylegan,shamshad2023clip2protect,kwon2022clipstyler}:
\begin{align}
\label{equation: visual quality loss}
    \mathcal{L}_{\mathrm{vis}}(\bm{w}) = \alpha_3 \cdot \mathcal{L}_{\mathrm{align}}(\bm{w}) + \alpha_4 \cdot \mathcal{L}_{\mathrm{latent}}(\bm{w}),
\end{align}
where $\alpha_3 \geq 0$ and $\alpha_4 \geq 0$ are hyperparameters that controls the trade-off. In particular, the first term ${\mathcal{L}_\mathrm{align}}$ in Equation \ref{equation: visual quality loss} captures the alignment of the embeddings regarding the text-image pairs between the initial and generated images based on a pre-trained CLIP model~\cite{radford2021learning}. 
Let $E_{\mathrm{I}}$ and $E_{\mathrm{T}}$ be the pre-trained CLIP image and text encoders, respectively, then $\mathcal{L}_{\mathrm{align}}$ is defined as:
\begin{align}
\label{equation: alignment loss}
    \mathcal{L}_\mathrm{align}(\bw) = 1 - \cos\big(\Delta I,  \Delta T \big),
\end{align}
where $\Delta I = E_{\mathrm{I}}(G_{\theta^*}(\bm{w})) - E_{\mathrm{I}}(G_{\theta^*}(\bm{w}_\mathrm{init}))$ is the distance between the image embeddings, $\Delta T = E_\mathrm{T}(p_{\mathrm{targ}}) - E_\mathrm{T}(p_{\mathrm{src}})$ denotes the distance between the text embeddings, 
and $p_{\mathrm{src}}$ is the semantic text prompt of the query image $\bm{x}_{\mathrm{ee}}$. 
Following \cite{shamshad2023clip2protect}, we simply set 
$p_{\mathrm{src}}$ as ``face'' since we are generally working with facial images.
Such an alignment loss promotes the generated images to follow the specifications of the target text prompt more closely. 
Finally, the last term $\mathcal{L}_{\mathrm{latent}}$ in Equation \ref{equation: visual quality loss} penalizes the Euclidean distance between the current latent code $\bm{w}$ and the initial latent code $\bm{w}_{\mathrm{init}}$:
\begin{align}
\label{equation: latent loss}
    \mathcal{L}_\mathrm{latent}(\bw) = \|\bm{w}-\bm{w}_\mathrm{init}\|_2.
\end{align}
By keeping $\bm{w}$ close to $\bm{w}_{\mathrm{init}}$, $\mathcal{L}_\mathrm{latent}$ ensures the visual similarity between initial and generated images.
Putting pieces together, we design the total loss function as $\mathcal{L}_\mathrm{tot}(\bm{w}) = \mathcal{L}_\mathrm{eff}(\bw) + \mathcal{L}_\mathrm{vis}(\bw)$ and perform gradient descent on $\mathcal{L}_{\mathrm{tot}}$ to obtain the optimized latent code $\bm{w}_S$ based on Equation \ref{equation: gradient descent}. Finally, $G_{\theta^*}(\bm{w}_S)$ will be returned as the protected Trackee image corresponding to the original $\bm{x}_{\mathrm{ee}}$.

\section{Experiments}
\label{section: experiments}

\begin{table*}[!t]
\centering
\caption{Comparisons of tracking success rates (\%) of static and dynamic FR strategies against different generative-based AFR methods across various image benchmarks and feature extractors. For each entry, the first number stands for TSR against the static FR strategy, and the second represents TSR against our DynTracker. Here, \textit{Clean} and \textit{Protected} mean Tracker starting with a single clean and protected Trackee image in the gallery database, respectively. For DynTracker, the lowest tracking success rate is bolded to highlight the best-performing AFR method in terms of Trackee's facial privacy protection.}
\vspace{-0.05in}
\resizebox{\textwidth}{!}{
\small
\centering
\begin{tabular}{l | l | c c | c c | c c | c c}
\toprule
\multirow{2.4}{*}{\textbf{Dataset}} & \multirow{2.4}{*}{\textbf{AFR Method}} & \multicolumn{2}{c|}{\textbf{MobileFace}} & \multicolumn{2}{c|}{\textbf{WebFace}} & \multicolumn{2}{c|}{\textbf{VGGFace}} & \multicolumn{2}{c}{\textbf{MagFace}} \\
\cmidrule{3-10}
& & \textbf{Clean} & \textbf{Protected} & \textbf{Clean} & \textbf{Protected} & \textbf{Clean} & \textbf{Protected} & \textbf{Clean} & \textbf{Protected} \\
\midrule
\multirow{7}{*}{FaceScrub}
& \small{No Protection} & $85.89 / 87.76$ & N/A & $98.17 / 98.29$ & N/A & $98.26 / 98.29$ & N/A & $99.29 / 99.36$ & N/A \\
\cmidrule{2-10}
& Adv-Makeup & $15.93 / 85.52$ & $45.93 / 82.62$ & $31.54 / 91.02$ & $58.66 / 88.84$ & $27.96 / 92.58$ & $59.42 / 89.16$ & $43.49 / 98.02$ & $62.34 / 97.12$ \\ 
& AMT-GAN & $13.38 / 85.42$ & $24.26 / 82.64$ & $29.87 / 92.35$ & $56.64 / 91.17$ & $30.29 / 92.58$ & $57.18 / 91.26$ & $44.25 / 99.87$ & $83.50 / 98.85$ \\ 
& Clip2Protect & $2.46 / 83.33$ & $31.37 / 83.33$ & $4.68 / 91.40$ & $42.64 / 89.65$ & $4.32 / 90.87$ & $43.85 / 88.60$ & $33.83 / 98.04$ & $80.39 / 98.04$ \\
& DiffAM & $14.54 / 84.96$ & $34.65 / 85.20$ & $27.64 / 92.72$ & $52.50 / 91.46$ & $24.86 / 91.25$ & $53.90 / 90.82$ & $39.65 / 98.26$ & $83.74 / 98.06$ \\
& \textbf{DivTrackee} & $2.67 / \mathbf{26.58}$ & $1.86 / \mathbf{22.16}$ & $7.82 / \mathbf{30.62}$ & $2.37 / \mathbf{29.49}$ & $5.94 / \mathbf{31.69}$ & $1.98 / \mathbf{32.68}$ & $6.82 / \mathbf{31.94}$ & $2.53 / \mathbf{28.57}$ \\
\midrule
\multirow{7}{*}{PubFig}
& No Protection & $86.29 / 88.07$ & N/A & $96.85 / 97.08$ & N/A & $98.01 / 98.15$ & N/A & $99.15 / 99.24$ & N/A \\ 
\cmidrule{2-10}
& Adv-Makeup & $24.37/83.38$ & $55.87/79.19$ & $34.46/86.12$ & $58.36/82.08$ & $32.18/88.54$ & $56.73/83.42$ & $35.68/94.57$ & $65.24/92.85$ \\ 
& AMT-GAN & $27.64/83.83$ & $56.44/79.67$ & $40.78/87.01$ & $49.64/84.82$ & $34.15/88.53$ & $51.07/85.44$ & $39.34/94.74$ & $60.85/93.16$ \\ 
& Clip2Protect & $3.45/81.93$ & $32.37/79.53$ & $4.83/84.69$ & $39.69/82.71$ & $5.35/86.28$ & $40.60/83.46$ & $42.63/93.70$ & $68.18/91.89$ \\
& DiffAM & $22.10 / 83.90$ & $52.46 / 80.40$ & $27.65 / 85.38$ & $46.09 /84.46$ &$28.24 / 87.32$ & $48.70 / 83.98$ & $38.10 / 94.65$ & $63.25 / 93.18$ \\
& \textbf{DivTrackee} & $3.23/\mathbf{28.57}$ & $2.88/\mathbf{29.58}$ & $4.82/\mathbf{32.17}$ & $2.63/\mathbf{29.72}$ & $5.25/\mathbf{35.76}$ & $3.27/\mathbf{32.54}$ & $5.88/\mathbf{37.06}$ & $4.71/\mathbf{34.53}$ \\
\midrule
\multirow{7}{*}{UMDFaces}
& No Protection & $86.89 / 87.42$ & N/A & $96.01 / 96.27$ & N/A & $98.19 / 98.33$ & N/A & $99.30 / 99.30$ & N/A \\
\cmidrule{2-10}
& Adv-Makeup & $18.93/84.28$ & $53.58/80.62$ & $29.23/87.38$ & $57.32/84.21$ & $25.03/88.77$ & $56.79/85.82$ & $33.68/96.08$ & $59.60/95.36$ \\ 
& AMT-GAN & $22.37/82.83$ & $47.68/80.60$ & $31.23/88.17$ & $57.72/83.83$ & $30.78/89.48$ & $56.28/84.25$ & $33.48/97.08$ & $61.34/94.88$ \\ 
& Clip2Protect & $3.01/81.57$ & $33.01/80.62$ & $4.54/88.61$ & $43.60/87.45$ & $5.38/90.18$ & $48.52/89.87$ & $31.46/95.48$ & $63.56/95.48$ \\ 
& DiffAM & $18.62 / 83.44$ & $45.18 / 81.36$ & $27.52 / 87.95$ & $51.46 / 87.02$ & $23.70 / 91.04$ & $52.26 / 91.32$ & $32.19 / 96.68$ & $60.46 / 96.30$\\
& \textbf{DivTrackee} & $3.69/\mathbf{29.17}$ & $3.22/\mathbf{27.54}$ & $5.63/\mathbf{31.64}$ & $3.89/\mathbf{28.79}$ & $4.86/\mathbf{32.61}$ & $3.46/\mathbf{29.57}$ & $5.18/\mathbf{33.64}$ & $4.10/\mathbf{31.15}$ \\
\midrule
\multirow{7}{*}{CelebA-HQ}
& No Protection & $82.30 / 85.83$ & N/A & $95.62 / 95.80$ & N/A & $96.90 / 97.08$ & N/A & $99.29 / 99.37$ & N/A \\ 
\cmidrule{2-10}
& Adv-Makeup & $42.38/90.26$ & $45.54/91.42$ & $46.68/91.39$ & $52.62/91.25$ & $47.70/92.40$ & $52.16/90.94$ & $49.85/90.62$ & $54.62/92.46$ \\ 
& AMT-GAN & $36.74/89.42$  & $38.36/90.50$  & $41.69/90.42$  & $46.78/92.40$  & $40.42/91.86$  & $43.60/90.74$  & $42.65/93.42$  & $44.55/91.84$  \\ 
& Clip2Protect & $16.34/86.62$ & $31.75/88.36$ & $19.42/92.38$ & $33.68/91.52$  & $21.40/91.98$  & $29.64/90.75$  & $22.56/94.64$  & $36.72/93.18$ \\ 
& DiffAM &$28.05 / 89.60$ & $36.42/ 88.16$ & $37.48 / 93.25$ & $48.20/92.66$ & $36.44/92.50$ &$43.84/91.94$ &$43.36/94.90$ & $46.72/93.56$ \\
& \textbf{DivTrackee} & $18.62/\mathbf{45.82}$ & $15.34/\mathbf{42.98}$ & $20.68/\mathbf{49.26}$ & $16.54/\mathbf{47.38}$ & $18.75/\mathbf{51.32}$ & $15.68/\mathbf{44.83}$ & $21.24/\mathbf{52.37}$ & $17.95/\mathbf{50.60}$ \\
\bottomrule
\end{tabular}
}
\label{table: main results}
\end{table*}

To systematically study the strength of our DynTracker and validate the superiority of our DivTracker regarding state-of-the-art AFR protection schemes, we conduct comprehensive experiments to compare the protection efficacy and visual aspects of different AFR methods across various image benchmarks and facial recognition settings (Section \ref{subsection: main results}). Besides, we also test the generalizability of our methods to facial verification models (Section \ref{subsection: application in facial verification}).
Below, we start with explaining the main setup of our experiments, while more detailed descriptions are provided in Appendix \ref{appendix: detailed experimental settings}.

\shortsection{Dataset}
We consider $4$ widely-used facial image benchmarks,  FaceScrub~\cite{ng2014data}, PubFig~\cite{kumar2009attribute}, UMDFaces~\cite{bansal2017umdfaces} and CelabA-HQ~\cite{karras2017progressive}, among which CelabA-HQ is the dataset with high-resolution facial images. 
All these datasets consist of images with annotated identities, each associated with multiple facial images. 

\shortsection{Configuration} For each benchmark dataset, we first randomly select $5$ identities as the target identities of Trackees. Then, for each Trackee identity, we generate AFR protections for all the Trackee images, including one image in the gallery database owned by Tracker, while treating the remaining images as the query images of Trackees. 
We follow Lowkey \cite{cherepanova2021lowkey} to set up the query and gallery images for the remaining identities other than Trackee.
We compare our DivTrackee with state-of-the-art generative-based methods, such as Adv-Makeup~\cite{yin2021adv}, AMT-GAN~\cite{hu2022protecting}, Clip2Protect~\cite{shamshad2023clip2protect} and DiffAM~\cite{sun2024diffam}.
For DivTrackee, we use the first $2000$ images from LFW~\cite{huang2008labeled} to construct the auxiliary dataset $\mathcal{X}_{\mathrm{aux}}$ in the diverse guidance loss and consider the text prompt $p_{\mathrm{targ}}$ as ``natural makeup''. 
We choose the hyperparameters involved in DivTrackee as follows: $m=10$, $\alpha_1 = 0.6$, $\alpha_2 = 1.2$, $\alpha_3 = 0.5$, $\alpha_4 = 0.02$, $\delta=0.2$, $\lambda=0.01$, and $S=60$ (see Section \ref{subsection: sensitivity hyperparameters} for our sensitivity analysis experiments). 

\shortsection{Feature Extractor}
To set the FR model of Tracker, we consider $4$ pre-trained facial feature extractors, including MobileFace~\cite{chen2018mobilefacenets},
WebFace~\cite{yi2014learning}, 
VGGFace~\cite{cao2018vggface2}, and
MagFace~\cite{meng2021magface}. 
To generate AFR protections for Trackee images, we use IR-50~\cite{hu2018squeeze}, IR-152~\cite{deng2019arcface}, and FaceNet~\cite{schroff2015facenet} as the substitute feature extractors to compute the cosine dissimilarity function $\cosDis_f$ for the adversarial losses in Equation \ref{equation: protection efficacy loss}.
Aligned with existing literature~\cite{hu2022protecting,shamshad2023clip2protect,sun2024diffam}, we consider the scenarios where feature extractors used for AFR generation differ from those employed by Tracker.
This black-box transferability setting simulates the scenario of having imprecise knowledge of Tracker's FR model when generating AFR protections.
 
\shortsection{Evaluation Metric}
We adopt \textit{tracking success rate} (TSR) to evaluate the strength of a specific Tracker's FR strategy. Specifically, TSR is defined as the ratio of the number of query images correctly identified as Trackee to the total number of Trackee images in the query dataset. The higher the TSR is, the stronger the employed FR strategy is. 
For Trackee, we use \emph{protection success rate} (PSR), defined as $1 - \text{TSR}$, to evaluate the efficacy of an AFR protection scheme.
In addition, we visualize the generated AFR-protected images to provide qualitative comparison results, which is considered the primary criterion for evaluating the visual quality of generated images (see Appendix \ref{appendix: quantitative comparison results} for additional quantitative comparisons).

\subsection{Main Results on Facial Recognition}
\label{subsection: main results}

Table~\ref{table: main results} summarizes the tracking success rates of Tracker's FR strategies against various generative-based AFR methods. We report the average tracking success rate for each AFR scheme over the $5$ selected Trackee identities against static FR strategy and our DynTracker, respectively.
We consider two settings of Tracker's initial knowledge of Trackee, either including a clean or an AFR-protected image in the gallery database gathered by Tracker.
As a reference, we evaluate the performance of tracking strategies when no AFR protection is employed on the set of Trackee query images, depicted as \textit{No Protection} in Table \ref{table: main results}.
For completeness, we also evaluate the performance of existing adversarial noise-based AFR protection schemes and provide the results in Table \ref{table: main results full} in Appendix \ref{appendix: full table results}.

\shortsection{Strength of DynTracker}
We start by looking at the tracking success rates of DynTracker when Trackee's protected images are crafted using existing AFR techniques. Table \ref{table: main results} shows that our dynamic FR strategy, DynTracker, is very effective in breaking the protections crafted by existing AFR generative-based AFR methods that do not promote diversity explicitly, achieving more than $80\%$ tracking success rates in almost all the tested scenarios. 
For example, DiffAM can only achieve protection success rates of around $10\%$ on the high-resolution CelebA-HQ dataset against our dynamic FR strategy.
Similar trends can be observed in Table \ref{table: main results full} for traditional adversarial noise-based AFR.
Compared with the scenario where no protections are deployed, existing AFR methods can only achieve a marginal improvement in protection success against DynTraker, confirming their limitations against determined trackers. 

In sharp contrast, the tracking success rates of static FR strategy are found to be significantly lower than DynTracker, especially when Tracker holds an initial clean Trackee image. 
This is the typical evaluation setting adopted in the prior AFR literature.
Even if Tracker initially includes a single Trackee's protected image in the FR model's gallery database, the protection efficacy of existing AFR methods clearly drops in most cases.
For instance, Clip2Protect is very effective on the FaceScrub dataset against the static FR strategy with MobileFace and a clean Trackee image, while TSR increases from $2.46\%$ to $31.37\%$ when the initial Trackee image is changed from clean to protected.
These empirical results align with our insights and preliminary results discussed in Section \ref{section: dynamic strategy of facial recognition}, suggesting a crucial overlooked aspect in AFR evaluations.
That said, if we use the static FR strategy to evaluate and compare AFR methods, we will obtain a largely overestimated protection efficacy due to the strong assumptions imposed on Tracker's behavior of not updating the FR model's gallery database with newly identified Trackee's protected images. This overestimation can lead to wrong concluding arguments and a false sense of security for AFR protections against real-world determined Trackers. Thus, we strongly recommend the community consider dynamic FR strategies like DynTracker when evaluating the efficacy of AFR protection schemes.

\shortsection{Effectiveness of DivTrackee}
Next, we examine the protection efficacy of our DivTrackee. Table \ref{table: main results} shows that DivTrackee consistently improves protection success rates against DynTracker over existing generative-based AFR methods.
In particular, DivTrackee maintains tracking success rates of around $30\%$ against DynTracker across FaceScrub, PugFig, and UMDFaces. For the high-resolution CelebA-HQ image dataset, DivTrackee achieves around $50\%$ protection success rates against DynTracker, which significantly improves protection efficacy over existing AFR methods with PSR usually less than $10\%$.
As we have explained in Section \ref{section: promoting diversity protects facial privacy}, the crucial difference lies in the diverse guidance and the diversity-promoting losses involved in DivTrackee, reducing the similarity between AFR-protected Trackee images during generation, which is the reason why DivTrackee can be much more resilient to dynamic FR strategies.
In addition, looking at the TSRs of the static FR strategy, we note that the protection efficacy with respect to DivTrackee is even better when the initial Trackee gallery image is protected, which is in sharp contrast to the decreased PSRs for existing AFR methods.
Our evaluation results highlight the usefulness of explicit diversity promotion in AFR protections against Trackers who update the gallery database. 
We believe that our exploration of diversifying the AFR-protected images in DivTrackee can serve as an important initial step toward more effective AFR protection schemes against determined Trackers in real-world application scenarios.

\begin{figure}[!t]
\centering
\includegraphics[width=0.98\linewidth]{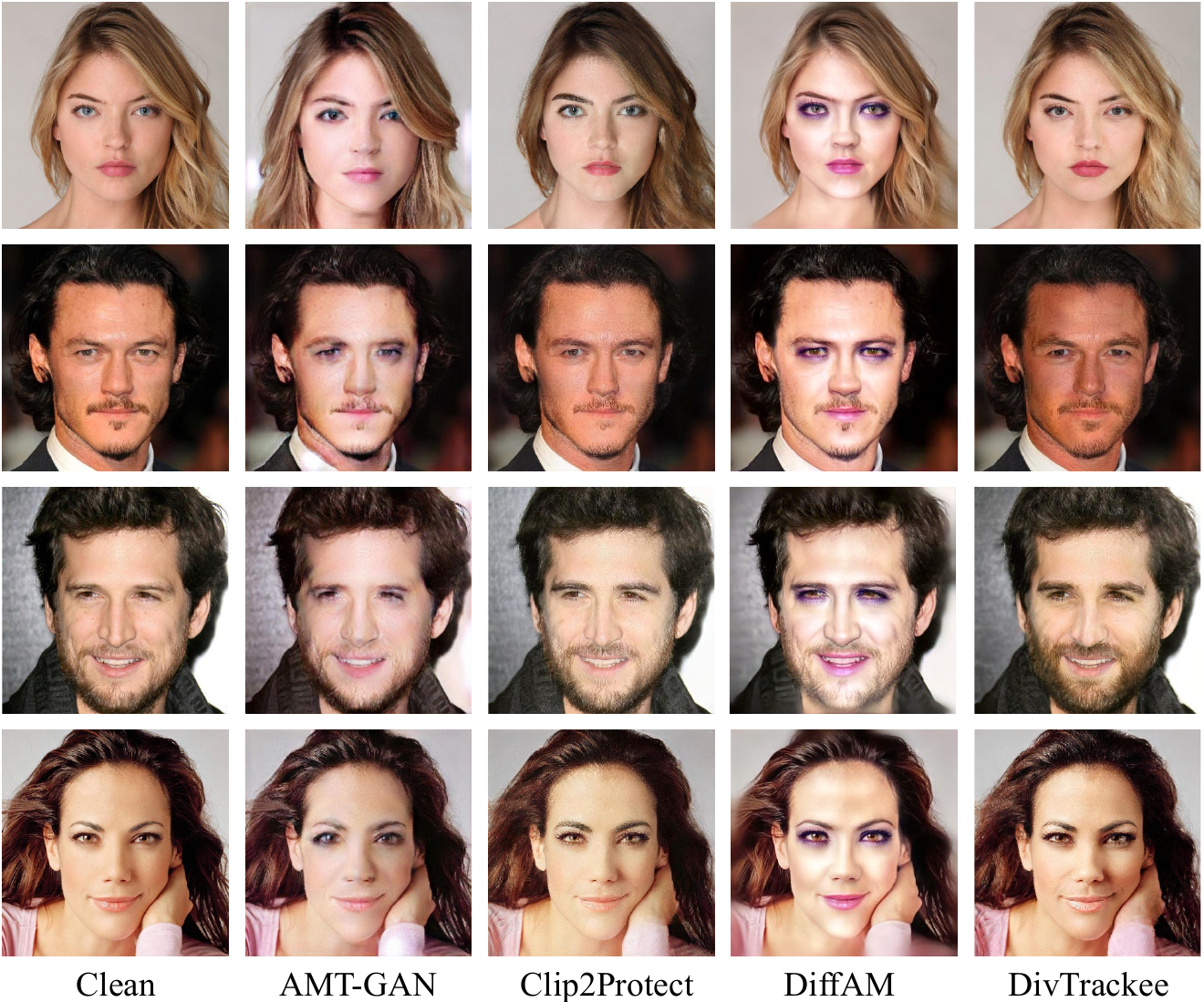}
\vspace{-0.1in}
\caption{Visualizations of protected Trackee images produced by different generative-based AFR methods. The target text prompt $p_{\mathrm{targ}}$ in DivTrackee is set as ``natural makeup''.}
\vspace{-0.1in}
\label{figure: visualization}
\end{figure}

\shortsection{Visual Quality}
Besides focusing on the primary goal of enhancing the protection efficacy of AFR, we also empirically study whether the protected images produced by DivTrackee can preserve the visual quality of the original facial images, which is a secondary but still important pursuit from Trackee's perspective.
In particular, we visualize the AFR-protected Trackee images generated by different query-target AFR techniques on high-resolution images from the CelebA-HQ dataset.

Figure~\ref{figure: visualization} compares DivTrackee with state-of-the-art generative-based AFR methods, including AMT-GAN, Clip2Protect, and DiffAM, to compare their capabilities of preserving the visual quality of original facial images.
We can observe that the protected facial images generated by {\ourDefense} look natural, without visual distortion of the original identity. 
In contrast, due to its severe impact at the pixel level, protected images generated by AMT-GAN can be easily distinguished from their clean version.
Despite preserving facial characteristics, Clip2Protect tends toward more sallow skin colors. DiffAM has a prominent makeup effect on the eyes and mouth, and the overall skin tone appears brighter, while protected images generated by DivTrackee are more vivid.
From the human perceptual view, {\ourDefense} can largely satisfy the expectations from Trackees for the visual quality pursuit of their images while maintaining strong protection performance.

\begin{figure*}[!t]
    \centering   \includegraphics[width=0.98\linewidth,height=4cm]{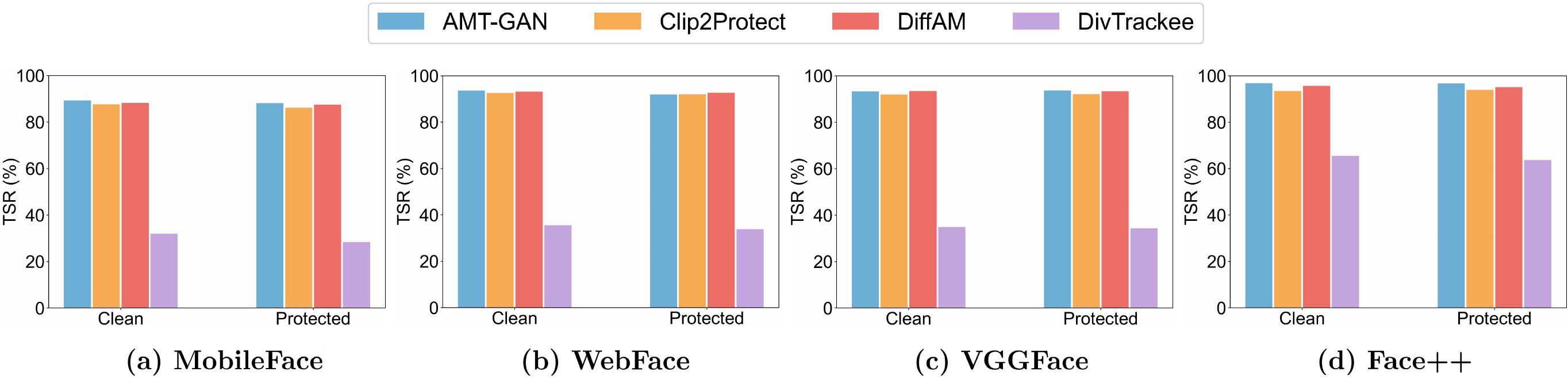}
    \vspace{-0.1in}
    \caption{Comparisons of tracking success rates (TSRs) of DynTracker adapted to facial verification settings against various generative-based AFR methods. Here, Figures (a)-(c) consider the scenarios where Tracker adopts some publicly available facial feature extractor, while a facial verification API, Face++, is tested in Figure (d).}
    \vspace{-0.05in}
    \label{fig: comparisons facial verification}
\end{figure*}

\begin{figure*}[!t]
    \centering   
    \begin{subfigure}[b]{0.24\linewidth}
        \includegraphics[width=\linewidth]{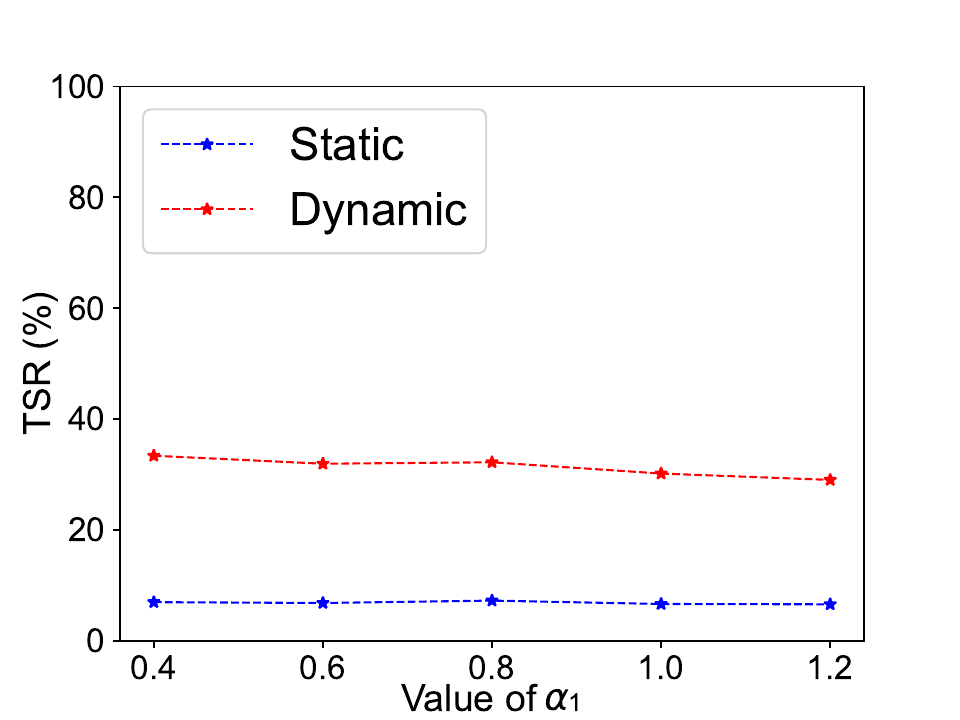}
        \caption{Impact of $\alpha_1$}
        \label{fig: sensitivity alpha 1}
    \end{subfigure}
    \begin{subfigure}[b]{0.24\linewidth}
        \includegraphics[width=\linewidth]{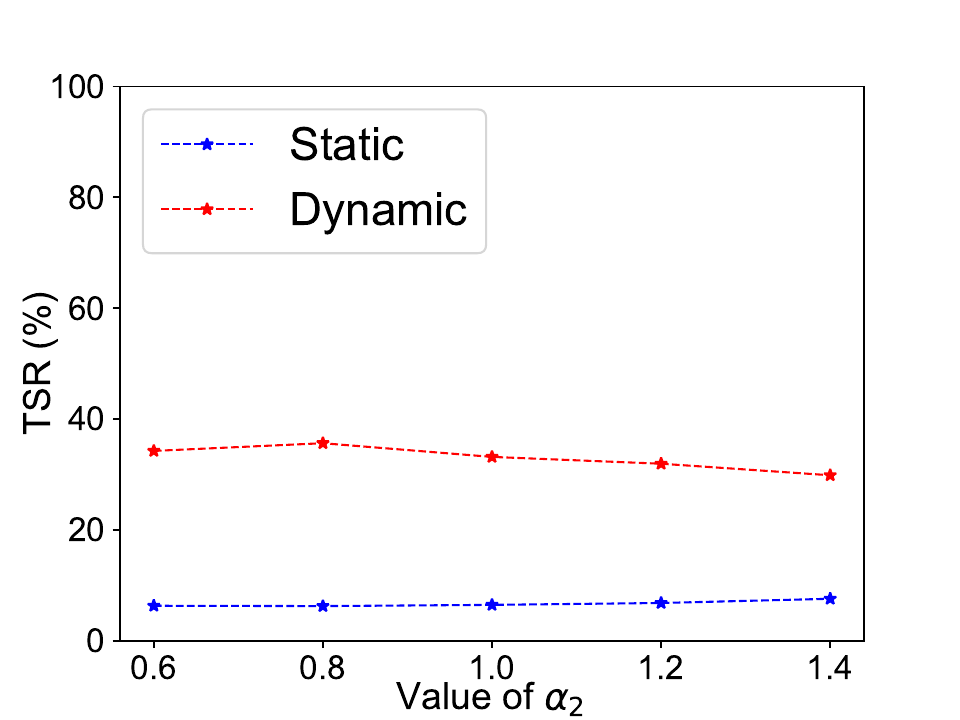}
        \caption{Impact of $\alpha_2$}
        \label{fig: sensitivity alpha 2}
    \end{subfigure}
    \begin{subfigure}[b]{0.24\linewidth}
        \includegraphics[width=\linewidth]{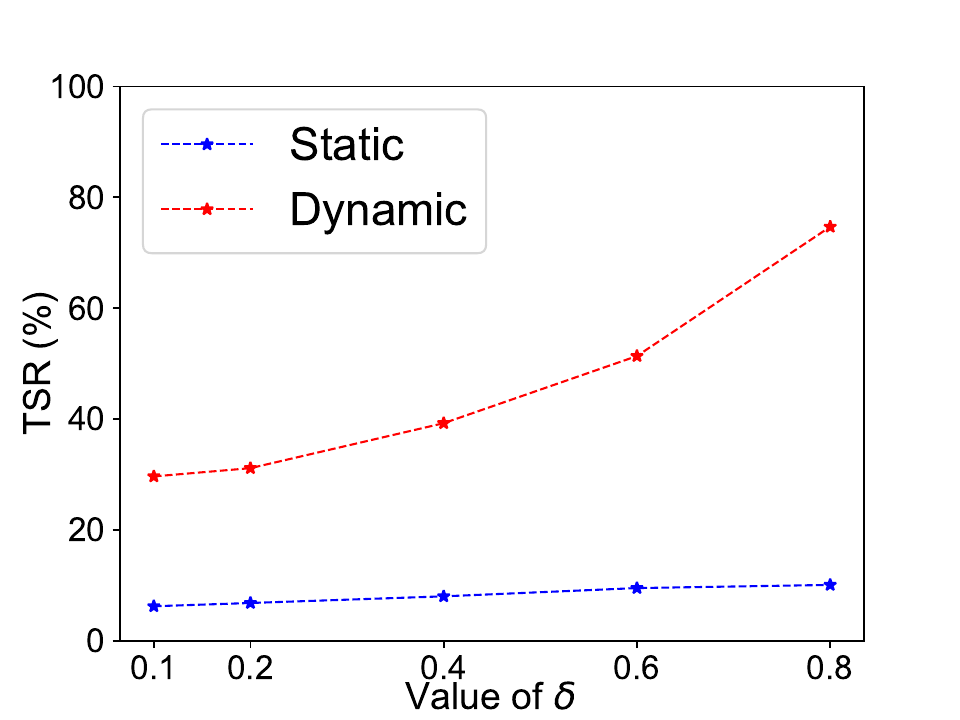}
        \caption{Impact of $\delta$}
        \label{fig: sensitivity delta}
    \end{subfigure}
    \begin{subfigure}[b]{0.24\linewidth}
        \includegraphics[width=\linewidth]{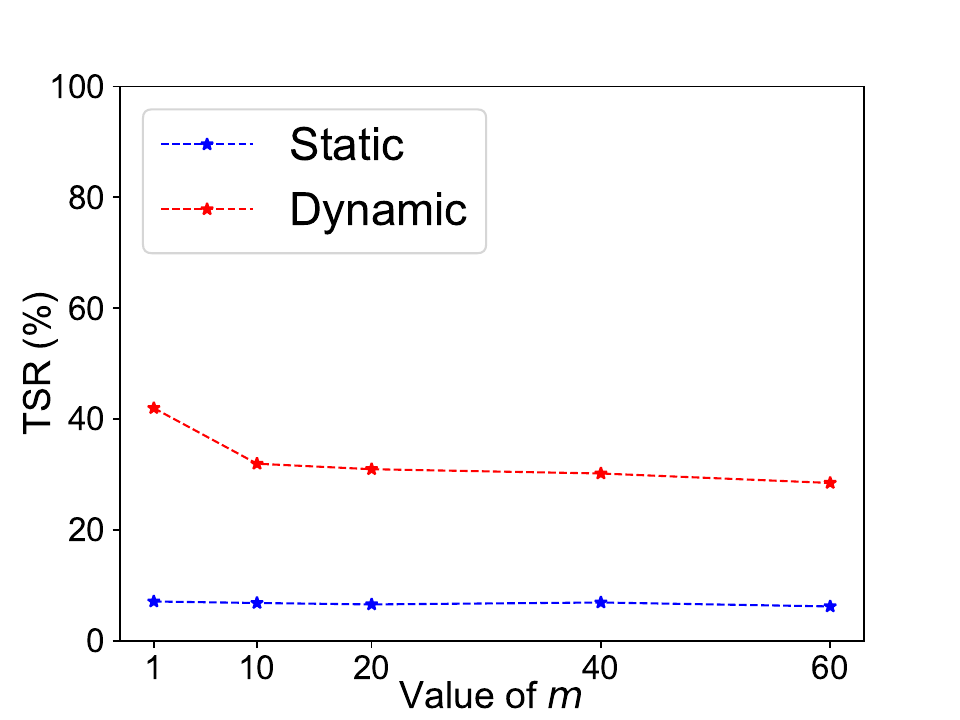}
        \caption{Impact of $m$}
        \label{fig: sensitivity m}
    \end{subfigure} 
    \vspace{-0.1in}
    \caption{Visualizations of sensitivity analysis results of key hyperparameters regarding the protection efficacy of DivTrackee.}
    \label{fig: sensitivity analysis}
    \vspace{-0.05in}
\end{figure*}

\subsection{Generalization to Facial Verification}
\label{subsection: application in facial verification}

Due to the similarity to facial recognition, Trackers have the freedom and capability to use a facial verification model to realize a similar goal of identifying Trackee images. 
Therefore, we further investigate whether the proposed DynTracker and DivTrackee can generalize to facial verification scenarios.
Unlike facial recognition models that require the construction of a gallery database, a reference image of a known identity (e.g., Trackee) and a predefined threshold $\gamma\in[-1,1]$ are sufficient to verify whether a given query image belongs to the reference identity.
Facial images from other individuals are not involved in the facial verification process.
Similar to the settings of DynTracker in Section \ref{section: dynamic strategy of facial recognition}, we assume Tracker initially holds a Trackee image, either clean or protected, as the reference image for facial verification.
During the dynamic updating stage, Tracker iteratively verifies the facial images in the query dataset based on the similarity scores to the reference image. 
As long as a successful match exists, the newly identified images will be regarded as reference images in the following iterations. 
When there are multiple reference images, a new query image is compared with each of them, which will be considered successful if any reference image yields a similarity score exceeding the threshold $\gamma$.
The tracking process will be terminated when no new query matches can be returned in the current iteration. 

\shortsection{Result}
Figure \ref{fig: comparisons facial verification} summarizes the comparisons of the tracking success rates of DynTracker adapted to facial verification across various generative-based AFR schemes. 
In this experiment, we consider the FaceScrub image dataset and different facial verification models, including $3$ publicly available feature extractors and Face++, a commercial facial verification API.\footnote{Face++ is available online at https://www.faceplusplus.com/.}
We set the threshold $\gamma$ to $0.542$, which is fixed throughout the tracking process. 
Similar to our facial recognition experiments, we report the tracking success rate averaged over the same $5$ selected identities to compare the efficacy of various AFR methods. 
Figure \ref{fig: comparisons facial verification} shows that DynTracker can be successfully applied to different facial verification frameworks, which all render the existing generative-based AFR methods ineffective. In sharp contrast, DivTrackee achieves significantly higher protection success rates against the facial verification adapted version of DynTracker.
For instance, DivTrackee lowers DynTracker's tracking success rates by more than $50\%$ compared to existing generative-based AFR methods.
These results again validate the strength of our dynamic tracking strategy and suggest that DivTrackee is superior in achieving high protection efficacy, particularly when the Tracker updates the reference Trackee images.  

\shortsection{Analysis} 
Compared with publicly available feature extractors, we observe a performance drop in the protection efficacy achieved by our DivTrackee when the Tracker employs the  Face++ API. 
We suspect that Face++
may have adopted data augmentation or adversarial training techniques to enhance the robustness of the employed facial feature extractor, making it easier
to identify the protected Trackee images generated by DivTrackee. This suggests that real-world APIs are powerful tools to implement DynTracker from Tracker's perspective, where developing better schemes to enhance the protection efficacy of DivTrackee against dynamic tracking strategy using API tools would be an interesting future direction.
In addition, we note that the TSRs of DynTracker under facial verification setup are slightly higher than those in Table \ref{table: main results}. 
Such a difference is likely due to the difference in the query matching step: a query match is regarded as successful in our facial verification experiments as long as there exists a reference image with similarity exceeding the threshold $\gamma$, whereas only the most similar gallery identity is returned for facial recognition.
As will be illustrated in Section \ref{subsection: false positive rate}, like what we observe under facial recognition, DynTracker under facial verification settings also induces only negligible false positive rates. Given the improved identification accuracy of Trackee images and negligible impact on false positive rates, dynamic tracking strategies with facial verification models pose significant threats to user facial privacy, where our diversity-promoting AFR method, DivTrackee, offers a viable solution to mitigate the privacy risks induced by DynTracker.

\section{Further Analysis and Discussion}
\label{section: additional analyses}

\begin{figure*}[!t]
    \centering   
    \begin{subfigure}[b]{0.25\linewidth}
        \includegraphics[width=\linewidth]{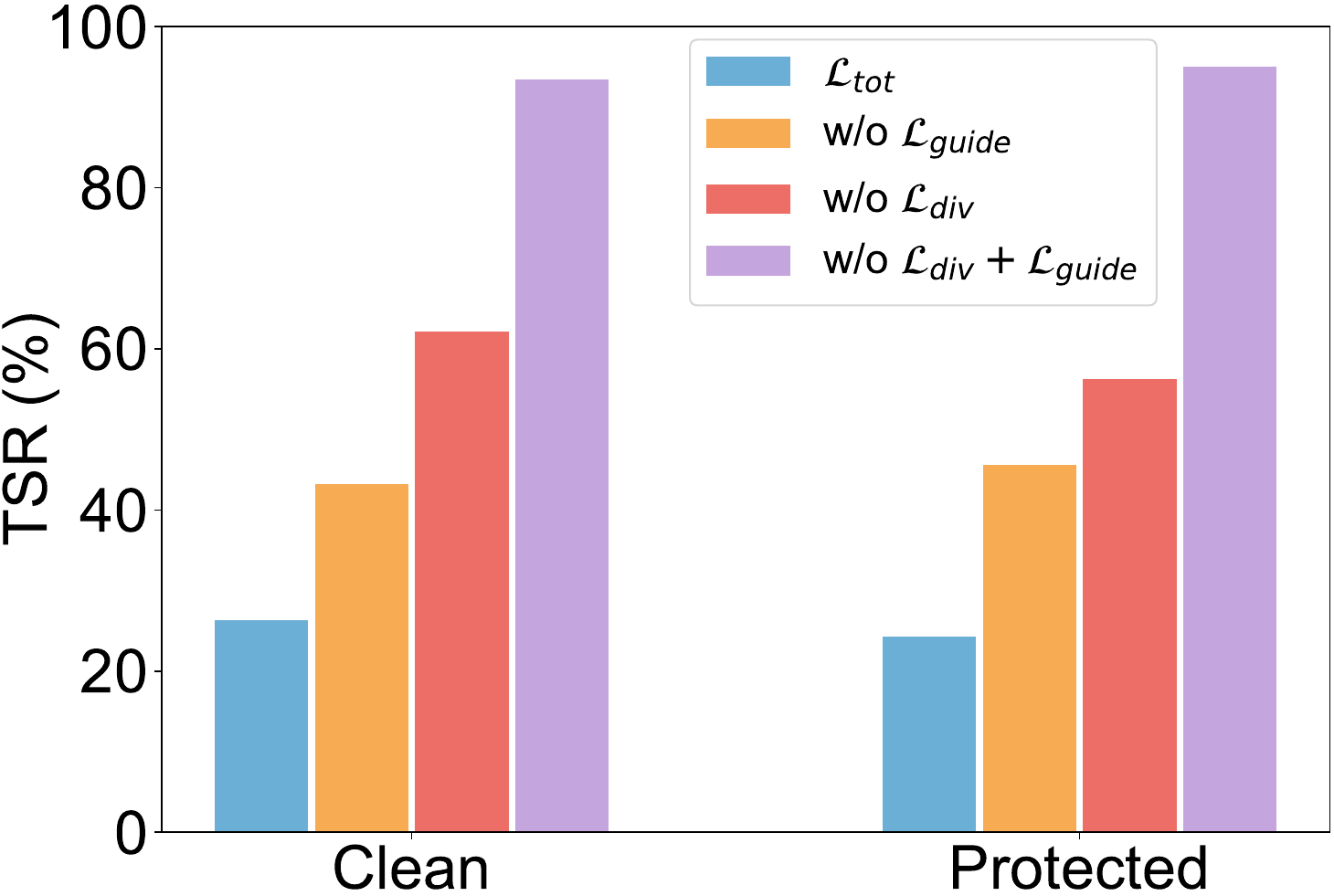}
        \caption{Facial Recognition}
        \label{figure: ablation study FR}
    \end{subfigure}
    \begin{subfigure}[b]{0.25\linewidth}
        \includegraphics[width=\linewidth]{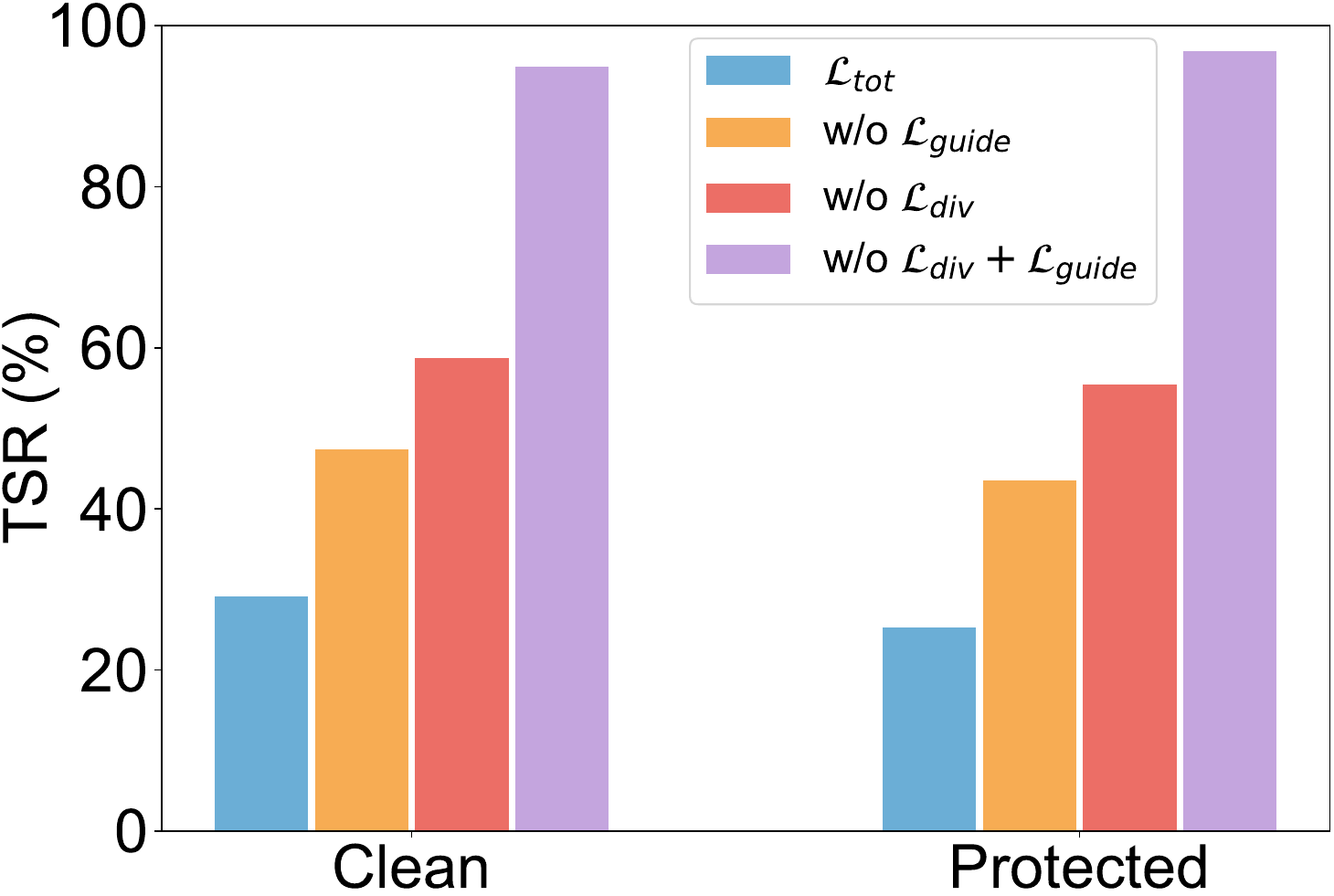}
        \caption{Facial Verification}
        \label{figure: ablation study FV}
    \end{subfigure}
    \hspace{0.02in}
    \begin{subfigure}[b]{0.23\linewidth}
        \centering
        \includegraphics[width=\linewidth, height=2.95cm]{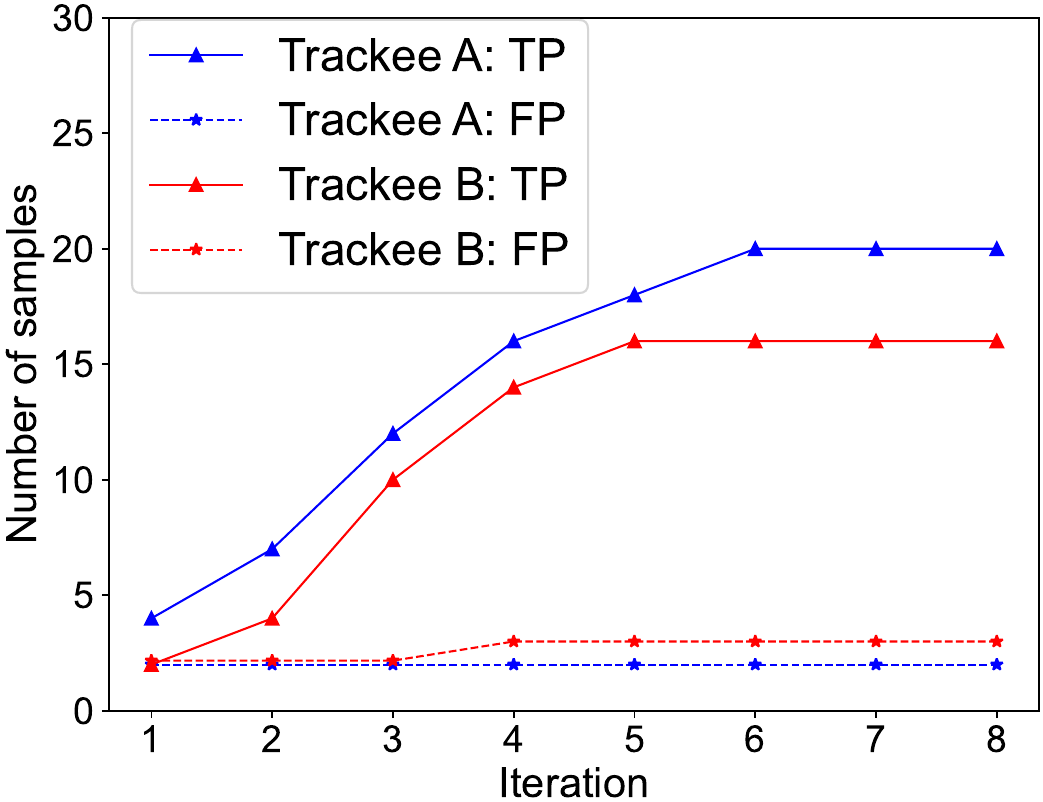}
        \caption{Facial Recognition}
        \label{figure: potential manual cost FR}
    \end{subfigure}
    \begin{subfigure}[b]{0.23\linewidth}
        \centering
        \includegraphics[width=\linewidth, height=2.95cm]{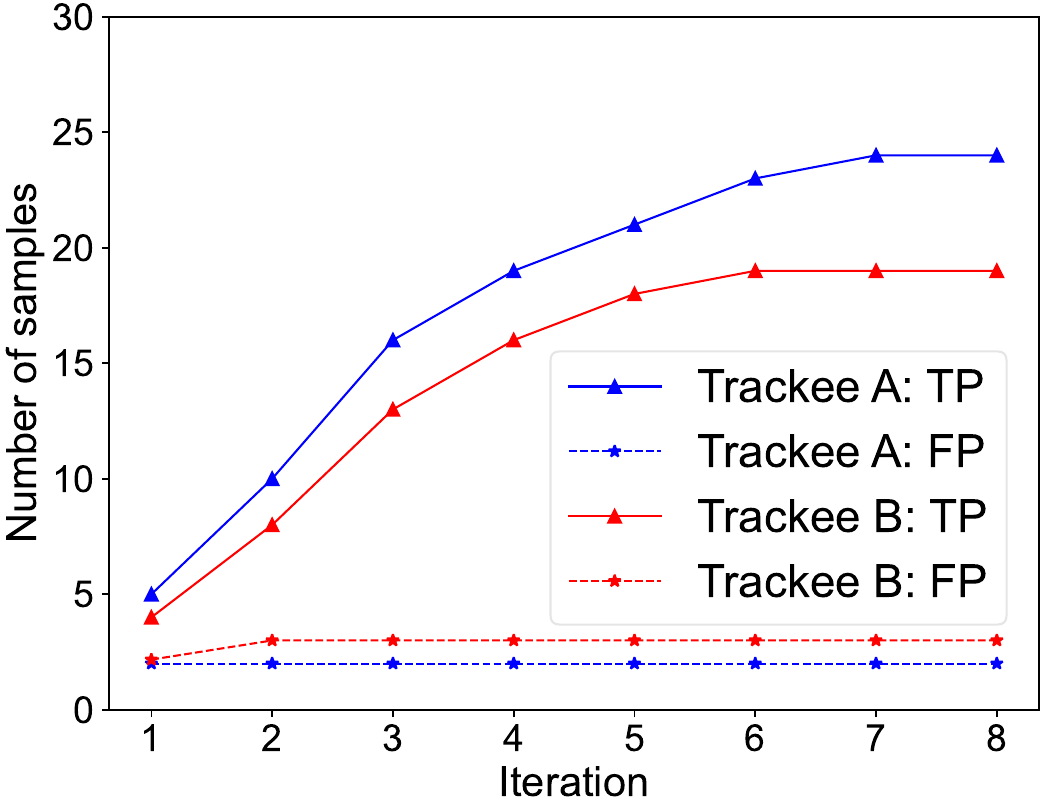}
        \caption{Facial Verification}
        \label{figure: potential manual cost FV}
    \end{subfigure}
    \vspace{-0.1in}
    \caption{(a)-(b) visualizes the effectiveness of the terms designed in our diversity-promoting adversarial loss, and (c)-(d) plots the curves of the number of true and false positives versus the number of dynamic updates in DynTracker.}
    \vspace{-0.05in}
\end{figure*}

In this section, we study the impact of hyperparameter choices and designed loss terms on the performance of DivTrackee (Sections \ref{subsection: sensitivity hyperparameters} and \ref{subsection: effectiveness of loss terms}), as well as how dynamically updating gallery database affects false positive rates (Section \ref{subsection: false positive rate}). All the experiments are conducted on the FaceScrub dataset without loss of generality.

\subsection{Hyperparameter Sensitivity }
\label{subsection: sensitivity hyperparameters}

We conduct sensitivity analyses of the key hyperparameters used in DivTrackee. Specifically, we consider the scenario where Tracker initially holds a clean Trackee image and employs MagFace as the facial feature extractor. We record the tracking success rates of DynTracker by varying the values of $\alpha_1, \alpha_2, \delta$, and $m$, where we observe similar trends for other experimental settings.
Figure \ref{fig: sensitivity analysis} visualizes our sensitivity analysis results, showing that DivTrackee is highly sensitive to the threshold parameter $\delta$ involved in $\mathcal{L}_{\mathrm{guide}}$, i.e., the TSR of DynTracker against DivTrackee sharply increases as we increase the value of $\delta$. 
In contrast, other hyperparameters have more minor impacts on DivTrackee's efficacy, and increasing their values decreases DynTracker's TSR slightly.
Also, we note that all the tested hyperparameters have almost no impact on the tracking success rates of the static FR strategy. 
This phenomenon aligns with our expectations, as these hyperparameters are associated with the diversity-promoting loss terms in $\mathcal{L}_{\mathrm{tot}}$ and thus are expected to be more influential to the protection success rates of DivTrackee when DynTracker is employed.
We chose the values of these hyperparameters for DivTrackee based on our sensitivity analyses, considering their impact on image generation quality.

\subsection{Design Choice of DivTrackee}
\label{subsection: effectiveness of loss terms}

To validate the key design choices of DivTrackee, we conduct ablation studies to examine the effectiveness of the proposed diversity-promoting adversarial loss terms in Equation \ref{equation: protection efficacy loss}.
We are particularly interested in how the diverse guidance loss $\mathcal{L}_{\mathrm{guide}}$ and the diversity-promoting loss $\mathcal{L}_{\mathrm{div}}$ affect the protection success rates of DivTrackee against DynTracker. 
Figures \ref{figure: ablation study FR} and \ref{figure: ablation study FV} illustrate the results under facial recognition and verification, respectively, where Tracker employs MagFace as the feature extractor of the FR model and involves an initial clean or protected Trackee image in the gallery database. 
To be more concrete, we compare the tracking success rates of {\ourAttack} when $\mathcal{L}_\mathrm{guide}$, $\mathcal{L}_\mathrm{div}$ or both are separately excluded from the total loss function used in DivTrackee for optimizing the latent code in Equation \ref{equation: gradient descent}.
We set the same remaining loss terms and choose the same hyperparameters for DivTrackee.

\shortsection{Result Analysis}
The blue bars in Figures \ref{figure: ablation study FR} and \ref{figure: ablation study FV} correspond to the performance of our DivTrackee with the total loss employed, where DynTracker exhibits the lowest tracking success rates.
When $\mathcal{L}_{\mathrm{guide}}$ is solely excluded from $\mathcal{L}_\mathrm{tot}$, the TSR of DynTracker increases by approximately $15\%$, which suggests the effectiveness of diversifying the auxiliary image guidance with a properly selected thresholding parameter $\delta$.
In addition, when we exclude $\mathcal{L}_\mathrm{div}$ from the total loss function, the tracking success rate increases more drastically with nearly $30\%$ on average. In comparison, the TSR of DynTracker can be as high as $95\%$ if both $\mathcal{L}_\mathrm{guide}$ and $\mathcal{L}_\mathrm{div}$ loss terms are ignored.
Our results highlight the effectiveness of the proposed losses in DivTrackee against determined Trackers, confirming the importance of explicitly promoting diversity in the design of AFR.

\subsection{Impact on False Positive Rate}
\label{subsection: false positive rate}

Although we expect well-trained FR models to be capable of accurately recognizing the identity from clean facial images, it is theoretically possible for clean query images of other identities to be misclassified as Trackee. 
If the employed FR model induces many false positive identifications, the Tracker will need to manually verify the recognized query images, which is particularly undesirable for DynTracker since it involves multiple rounds of facial recognition.
Therefore, we conduct experiments to investigate the impact of false positives during the tracking process of DynTracker.
Specifically, we first adopt a $7/3$ split on FaceScrub to construct training and testing sets and then record the testing identities into which clean images of other individuals are misclassified. Two such identities are selected as Trackee, denoted as Trackee A and Trackee B.
Figures \ref{figure: potential manual cost FR} and \ref{figure: potential manual cost FV} visualize how the number of true and false positives changes during the dynamic update stage of DynTracker, where we consider both facial recognition and verification settings. Similar to our experiments in Section \ref{subsection: sensitivity hyperparameters}, Tracker is assumed to adopt MagFace and initially hold a clean Trackee image. We assume Trackee employs our DivTrackee as the AFR protection scheme.

\shortsection{Result Analysis}
According to our experimental design, both selected Trackees incur two false positives at the initial iteration of DynTracker. 
Note that these false positives will be included and paired with the Trackee identity in the FR model's gallery database after the first iteration of DynTracker.
However, as the gallery database is updated with newly identified Trackee query images, surprisingly, $0$ (resp. $1$) additional false positives are introduced for Trackee A (resp. Trackee B) during the subsequent tracking process. 
In comparison, the number of true positives for both Trackees steadily increases until it plateaus.
These results not only confirm the effectiveness of dynamically updating the gallery database in boosting the tracking success rate but also validate the negligible impact of DynTracker on false positive rates.
One possible explanation for the negligible increase in FPR is that since the most similar gallery identity is always returned and the gallery database contains many images, it is unlikely that the query matching step picks the false positive samples included in the gallery database.

\subsection{Discussion and Future Work} 

\shortsection{Computation}
For DynTracker, even when the gallery database contains over $40,000$ images, the employed FR model can still quickly predict the identity of a query facial image in an average of $28$ms. Each update of the nearest neighbor classifier generated from the gallery database takes an average of $8$ms. The dynamic update stage of DynTracker typically takes $4$-$8$ iterations to complete.
Since our DynTracker is not intended for real-time prediction, we regard the additionally-incurred computational costs as negligible from Tracker's perspective.
In addition, the most time-consuming step for DivTrackee is generator fine-tuning, which is designed to improve the generator's generalizability to individual Trackee facial images but often takes $1$-$2$ minutes.
Exploring how to reduce the computation overhead or even avoid the costly generative fine-tuning step in AFR could be an interesting future direction.

\shortsection{Effectiveness} 
Despite achieving high PSRs on FaceScrub, PubFig, and UMDFaces, DivTrackee's performance degrades on the CelebA-HQ dataset against dynamic FR strategies. This may be because high-resolution images contain more facial feature details, and achieving adversarial effects requires adding more perturbations, thereby impacting the visual quality of the generated AFR-protected images.
Future work can explore ways to improve DivTrackee's protection efficacy on high-resolution datasets.
In addition, this paper does not consider potential time shifts in Trackee's facial images, such as the differences in appearance between childhood and adulthood. 
It remains an open question whether DynTracker can effectively track recently posted Trackee images when the initial gallery image is collected from much earlier periods.

\begin{table}[!t]
\centering
\caption{TSR comparisons of DynTracker with various data preprocessing schemes against DivTrackee on FaceScrub.}
\vspace{-0.1in}
\resizebox{0.85\columnwidth}{!}{
\begin{tabular}{l | c c}
    \toprule
    \textbf{Preprocessing Scheme} & \textbf{Clean}& \textbf{Protected} \\
    \midrule
    No Preprocessing & $31.94\%$ & $28.57\%$ \\
    \midrule
    JPEG Compression & $32.87\%$ & $29.16\%$ \\
    Gaussian Smoothing & $34.52\%$ & $36.10\%$ \\
    \bottomrule
\end{tabular}
}
\vspace{-0.1in}
\label{table: preprocess}
\end{table}

\shortsection{Image Preprocessing} Tracker might adopt state-of-the-art facial recognition models or more robust models in the future, or apply preprocessing techniques (such as image compression or makeup removal tools) to compromise the adversarial perturbations added to the Trackee facial images. Our experiments in Section \ref{subsection: application in facial verification} suggest that when DynTracker is implemented based on a more robust FR model (e.g., the Face++ facial verification API), the protection efficacy of DivTrackee can deteriorate.
As a preliminary exploration, we conduct experiments to examine the sensitivity of DivTrackee's performance to two simple data preprocessing schemes, including JPEG compression and Gaussian Smoothing.
These preprocessing steps were also studied in existing literature \cite{cherepanova2021lowkey}.
The results are demonstrated in Table~\ref{table: preprocess}, where we consider the FaceScrub dataset and implement our DynTracker based on the MagFace feature extractor.
We can observe that DynTracker's tracking success rate slightly improves when Trackee images are preprocessed before feeding them into the FR model. 
Nevertheless, Tracker needs to ensure that the preprocessing methods do not remove excessive facial features of the target identity; otherwise, integrating preprocessing techniques might adversely affect DynTracker's tracking success. 

\shortsection{Adaptive Variations of FR Tracking Strategies}
Our proposed DynTracker can be regarded as a specific adaptive FR strategy that is likely to be adopted by determined trackers.
However, there are also other variants of tracking strategies that are worth discussing.
In this work, Tracker is assumed to include a single Trackee facial image, either clean or protected, in the gallery database to launch DynTracker.
In reality, however, Tracker may have already collected multiple clean or protected images by following Trackee's social media accounts.
Including more Trackee images might increase DynTracker's tracking success rates, aligned with the incentives of determined trackers and real-world application scenarios. 
In addition, it is worth studying the scenarios in which the FR model's feature extractor is further fine-tuned using the additional recognized Trackee query images.
Note that the feature extractor is assumed to be fixed in DynTracker to simplify the tracking strategy.
Similar to the idea of DynTracker versus the static FR strategy in terms of the gallery database, dynamically updating the feature extractor using additional Trackee images may also improve tracking success, compared with fixing the feature extractor.
However, the challenge lies in the fact that Tracker does not know precisely about Trackee’s AFR strategy. Instead, Tracker needs to rely on a small amount of AFR-protected images to implement adaptive fine-tuning or unlearning, which is more difficult than building adversarial defenses against specific (known types of) perturbations.
Nevertheless, we consider exploring stronger adaptive modifications to DynTracker that may break our protections as interesting future work.
On the flip side, developing effective schemes based on our DivTrackee's insight to be robust to the aforementioned (adaptive) variations of FR tracking strategies would be an important future direction for advancing anti-facial recognition technology.

\shortsection{Game between Tracker and Trackee}
Again, we want to emphasize that in the security game between Tracker and Trackee (Definition \ref{definition: game between tracker and trackee}), Trackee is perpetually at a disadvantage since Trackee cannot modify the facial images once they are published. In contrast, Tracker can iteratively refine the tracking strategy, aiming to identify more Trackee posted images~\cite{radiya-dixit2022data}. Such asymmetry does not imply that the Trackee is at a ``fundamental'' disadvantage but rather represents a challenging task that the community can work on to find better solutions. At a higher level, such asymmetry between Tracker and Trackee is similar to the problem of building robust models against adversarial perturbations, where the defender (like Trackee) is at a disadvantage compared with the attacker (like Tracker). The fundamental difference lies in that we do not have a clear picture of how to model Tracker's behavior, as opposed to the explicit constraint that we usually impose on adversarial examples. Our work makes an initial step by introducing the precise working pipeline of DynTracker, which is strictly more potent than its static counterpart, relies only on necessary assumptions, and is easy to implement. This enables us to characterize the actual behaviors of determined Trackers better and establish an essential foundation for developing more reliable AFR technologies.

\section{Conclusion}
\label{section: conclusion}

We designed a simple but highly effective dynamic FR strategy, DynTracker, where the model's gallery database is progressively updated with newly recognized target identity images. Our work reveals the limitations of existing AFR, particularly highlighting the importance of using dynamic FR strategies for more rigorous AFR evaluations. Besides, we develop DivTrackee, an innovative text-guided generative AFR method that leverages diversity-promoting modules and adversarial losses. It explicitly encourages more diverse generations of AFR-protected images as a promising solution to defend against DynTracker.
We hope that our comprehensive investigations of dynamic FR tracking strategies and their countermeasures can inspire more reliable designs of anti-facial recognition techniques that can potentially be deployed for practical use.

\section*{Availability}

Our DynTracker and DivTrackee implementations, as well as all our experiments, are available as open-source
code at \href{https://github.com/fiora6/divtrackee}{\textcolor{red}{this url}}.

\begin{acks}
This work is supported by the National Key R\&D Program of China under Grant 2022YFB3103500, the National Natural Science Foundation of China under Grant 62402087 and 62020106013, the Chengdu Science and Technology Program under Grant 2023-XT00-00002-GX, the Sichuan Science and Technology Program under Grant 2024ZHCG0188 and 2025ZNSFSC1490, the Fundamental Research Funds for Chinese Central Universities under Grant ZYGX2024J019, the China Postdoctoral Science Foundation under Grant BX20230060, BX20240053 and 2024M760356.
\end{acks}

\bibliographystyle{ACM-Reference-Format}
\bibliography{ref}

\clearpage
\newpage

\appendix
\onecolumn

\section{Algorithm Pseudocode}
\label{appendix: algorithm pseudocode}

\begin{algorithm}[H]
\caption{DynTracker: Dynamic Facial Recognition Strategy}
\label{algorithm: DynTracker}
\setstretch{1.1}
\begin{algorithmic}[1]
\Function{DynTracker}{facial recognition model $\mathrm{FR}$, initial gallery database $\galleryDatabase$, query images $\tilde{\queryImages}$}
    \State Initialize \ $t \gets 0$, \ $\TrackeeImages \gets \{\}$, \  $\galleryDatabase(0) \gets \galleryDatabase$ \ and \ $\tilde{\queryImages}(0) \gets \tilde{\queryImages}$
    \While {True}
        \State $t \gets t+1$
        \State $\TrackeeImages(t) \gets \big\{ \bm{x}_{\q}\in\tilde\queryImages(t-1) \ \big| \ \mathrm{FR} \left( \bm{x}_{\q}; \galleryDatabase(t-1) \right) = y_{\mathrm{ee}} \big\}$ \Comment{Recognize Trackee query images}
        \State $\galleryDatabase(t) \gets \galleryDatabase(t-1)\cup \left\{(\bx, y_{\mathrm{ee}}) \ \big| \ \bx \in \TrackeeImages(t) \right\}$ \Comment{Update FR model's gallery database}
        \State $\tilde\queryImages(t) \gets \tilde\queryImages(t-1) \setminus \TrackeeImages(t)$
        \vspace{0.1cm}
        \If {$|\TrackeeImages(t)| > 0$}
            \State $\TrackeeImages \gets \TrackeeImages\bigcup\TrackeeImages(t)$ \Comment{Merge recognized Trackee query images}
        \Else
            \State Break 
            \Comment{Terminate if no more images can be recognized}
        \EndIf
    \EndWhile
    \State \Return $\TrackeeImages$
\EndFunction
\end{algorithmic}
\end{algorithm}

\begin{algorithm}[H]
\caption{DivTrackee: Diversity-Promoting Anti-Facial Recognition}
\label{algorithm: DivTrackee}
\setstretch{1.1}
\begin{algorithmic}[1]
\Function{DivTrackee}{generator $G_{\theta}$, clean Trackee image $\bm{x}_\mathrm{ee}$, auxiliary dataset $\mathcal{X}_{\mathrm{aux}}$, target text prompt $p_{\mathrm{targ}}$, dissimilarity function $\cosDis_f$, e4e encoder $F_\phi$, CLIP image encoder $E_{\mathrm{I}}$, CLIP text encoder $E_{\mathrm{T}}$, current queue $\mathcal{Q}$, hyperparameters $\alpha_1,\alpha_2,\alpha_3,\alpha_4, \delta, m, \lambda, S$}
    \State $G_{\theta^*} \gets \mathrm{FineTune}(G_\theta, F_\phi, \bm{x}_\mathrm{ee})$ 
    \Comment{Fine-tune the StyleGAN generator according to Appendix \ref{appendix: generator fine-tuning}}
    \State Initialize \ $\bm{w}_{\mathrm{init}} \gets F_\phi(\bm{x}_\mathrm{ee})$ \ and \ $\bm{w}_0 \gets \bm{w}_\mathrm{init}$ 
    \Comment{Initialize the latent code based on GAN inversion}
    \For{$s  = 0, 1, 2, \ldots, S-1$}
        \State \emph{// Compute diversity-promoting adversarial loss terms in Equation \ref{equation: protection efficacy loss}}
        \State $\bm{x}_{\mathrm{aux}} \gets$ Randomly select an image from $\mathcal{X}_{\mathrm{aux}}$
        \State $\mathcal{L}_{\mathrm{adv}}(\bw_s) = -\cosDis_f \big( G_{\theta^*}(\bw_s), \bm{x}_{\mathrm{ee}} \big)$
        \State $\mathcal{L}_{\mathrm{guide}}(\bw_s) = \max \big( 0,D_f \left( G_{\theta^*}(\bw_s),\bx_{\mathrm{aux}} \right) -\delta \big)$
        \If {$|\mathcal{Q}| > 0$}
            \State $\mathcal{L}_{\mathrm{div}}(\bw_s) = - \frac{1}{m} \sum_{\bm{x}\in\mathcal{Q}} \ D_f\big(G_{\theta^*}(\bw_s), \bm{x}\big)$
            \Comment{Note that if $|\mathcal{Q}| = 0$, $\mathcal{L}_{\mathrm{div}} = 0$}
        \EndIf
        \State \emph{// Compute visual quality loss terms in Equation \ref{equation: visual quality loss}}
        \State $\mathcal{L}_\mathrm{align}(\bw_s) = 1 - \cos\big( \ E_{\mathrm{I}}(G_{\theta^*}(\bw_s)) - E_{\mathrm{I}}(G_{\theta^*}(\bm{w}_\mathrm{init})), E_\mathrm{T}(p_{\mathrm{targ}}) - E_\mathrm{T}(p_{\mathrm{src}}) \ \big)$
        \Comment{Source text prompt $p_{\mathrm{src}}$ is set as ``face''}
        \State $\mathcal{L}_\mathrm{latent}(\bw_s) = \|\bw_s-\bm{w}_\mathrm{init}\|_2$
        \State \emph{// Optimize latent code using gradient descent based on the total loss}
        \State $\mathcal{L}_\mathrm{tot}(\bw_s) = \mathcal{L}_{\mathrm{adv}}(\bw_s) + \alpha_1 \cdot \mathcal{L}_{\mathrm{guide}}(\bw_s) + \alpha_2 \cdot \mathcal{L}_{\mathrm{div}}(\bw_s) + \alpha_3 \cdot \mathcal{L}_{\mathrm{align}}(\bw_s) + \alpha_4 \cdot \mathcal{L}_{\mathrm{latent}}(\bw_s)$ 
        \State $\bm{w}_{s+1} = \bw_s - \lambda\nabla_{\bm{w}}\mathcal{L}_\mathrm{tot}(\bw_s)$ 
    \EndFor
    \If {$|\mathcal{Q}|$ < m}
        \State Append $G_{\theta^*}(\bm{w}_S)$ to the end of $\mathcal{Q}$   
    \Else 
        \State Update $\mathcal{Q}$ with $G_{\theta^*}(\bm{w}_S)$ in a FIFO manner   
        \Comment{Ensure the length of $\mathcal{Q}$ not exceeding $m$ }
    \EndIf
    \State \Return $G_{\theta^*}(\bm{w}_S)$
\EndFunction
\end{algorithmic}
\end{algorithm}

\subsection{Generator Fine-Tuning}
\label{appendix: generator fine-tuning}

To achieve better editing performance for out-of-distribution (OOD) face images while maintaining image quality, we adopt the techniques proposed by~\cite{roich2022pivotal} to fine-tune the generator for each Trackee query image. More specifically, the generator fine-tuning step can be cast into an optimization problem with respect to the weight parameters of $G_\theta$ as follows:
\begin{align*}
    \theta^*= \argmin_{\theta} \ \mathcal{L}_{\mathrm{LPIPS}} \big( \bm{x}_{\mathrm{ee}}, G_\theta\left(
    \bm{w}_{\mathrm{init}}
    \right) \big) + \lambda \cdot \big\| \bm{x}_{\mathrm{ee}} - G_\theta \left( \bm{w}_{\mathrm{init}} \right) \big\|_2,
\end{align*}
where $\mathcal{L}_{\mathrm{LPIPS}}$ denotes the LPIPS perceptual loss, $\bm{w}_{\mathrm{init}}$ is the inverse latent code of Trackee's clean image $\bm{x}_{\mathrm{ee}}$, and $\lambda>0$ is a hyperparameter. In our experiments, we set $\lambda$ to $0.5$ and fine-tuned the generator for $450$ epochs.

\begin{table*}[t]
\centering
\caption{Comparisons on TSRs (\%) of static and dynamic FR strategies against various noise-based AFR methods. For each entry, the first number stands for TSR against the static FR strategy, and the second represents TSR against our DynTracker.}
\vspace{-0.1in}
\small
\resizebox{0.98\textwidth}{!}{
\centering
\begin{tabular}{l | l | c c | c c | c c | c c}
\toprule
\multirow{2.4}{*}{\textbf{Dataset}} & \multirow{2.4}{*}{\textbf{Method}}   
& \multicolumn{2}{c|}{\textbf{MobileFace}} & \multicolumn{2}{c|}{\textbf{WebFace}} & \multicolumn{2}{c|}{\textbf{VGGFace}} & \multicolumn{2}{c}{\textbf{MagFace}} \\
\cmidrule{3-10}
& & \textbf{Clean} & \textbf{Protected} & \textbf{Clean} & \textbf{Protected} & \textbf{Clean} & \textbf{Protected} & \textbf{Clean} & \textbf{Protected} \\
\midrule
\multirow{4}{*}{FaceScrub}
& PGD & $7.42/84.54$ & $30.22/81.69$ & $10.69/91.24$ & $38.84/90.35$ & $10.48/92.15$ & $39.62/90.24$ & $28.47/98.76$ & $48.32/97.66$ \\ 
& MI-FGSM & $6.15/83.68$ & $29.15/79.25$ & $9.35/90.28$ & $40.84/87.69$ & $10.06/91.41$ & $39.62/90.24$ & $31.93/98.94$ & $47.36/97.53$ \\ 
& TI-DIM & $4.62/83.26$ & $27.56/79.54$ & $6.76/90.22$ & $41.80/88.06$ & $11.34/90.23$ & $40.53/88.15$ & $25.57/97.80$ & $45.87/96.68$ \\ 
& TIP-IM & $1.98/83.96$ & $20.28/80.86$ & $4.26/91.38$ & $28.65/89.10$ & $5.13/91.81$ & $30.82/88.30$ & $13.82/97.10$ & $40.56/96.54$ \\ 
\midrule
\multirow{4}{*}{PubFig}
& PGD & $10.79/80.14$ & $29.97/77.12$ & $8.34/82.65$ & $35.71/79.56$ & $11.13/85.08$ & $39.25/80.93$ & $28.68/94.46$ & $52.38/93.34$ \\ 
& MI-FGSM & $8.63/81.27$ & $31.45/75.26$ & $8.42/85.82$ & $32.63/78.94$ & $9.56/87.13$ & $33.47/81.06$ & $29.78/94.37$ & $51.88/93.52$ \\ 
& TI-DIM & $6.87/81.69$ & $32.12/76.54$ & $7.25/82.53$ & $33.64/79.32$ & $10.18/85.06$ & $34.51/82.15$ & $21.73/94.43$ & $40.69/93.57$ \\ 
& TIP-IM & $3.07/81.89$ & $31.76/76.02$ & $5.18/80.28$ & $35.62/78.08$ & $6.13/83.04$ & $34.74/80.76$ & $13.48/94.11$ & $41.51/92.34$ \\ 
\midrule
\multirow{4}{*}{UMDFaces}
& PGD & $9.68/83.27$ & $32.58/79.97$ & $12.83/88.87$ & $38.65/85.66$ & $10.28/89.12$ & $39.82/86.32$ & $26.74/96.73$ & $49.74/95.42$ \\ 
& MI-FGSM & $7.87/82.76$ & $32.14/78.52$ & $9.48/87.37$ & $37.81/84.45$ & $13.77/88.63$ & $39.67/85.88$ & $21.43/96.13$ & $45.17/95.26$ \\ 
& TI-DIM & $6.68/82.87$ & $30.90/77.06$ & $7.78/85.07$ & $29.27/82.29$ & $9.87/87.23$ & $36.20/83.15$ & $18.78/95.34$ & $49.28/94.18$ \\ 
& TIP-IM & $2.78/79.37$ & $24.87/74.25$ & $5.77/86.23$ & $25.92/80.04$ & $3.78/87.54$ & $27.94/81.62$ & $15.87/93.34$ & $24.83/92.30$ \\ 
\midrule
\multirow{4}{*}{CelebA-HQ} 
& PGD & $24.37/89.60$  & $23.28/87.66$ & $26.10/90.62$ & $22.68/87.10$ & $29.60/89.65$ & $27.32/87.54$ & $31.02/91.37$ & $29.62/87.23$ \\ 
& MI-FGSM & $24.88/88.72$ & $22.16/86.48$ & $26.72/90.62$ & $26.90/91.82$ & $27.44/91.37$ & $25.92/90.69$ & $28.73/92.05$ & $24.80/91.60$  \\ 
& TI-DIM & $21.64/90.58$ & $23.62/91.35$ & $22.36/92.60$ & $25.42/90.82$ & $23.70/90.85$ & $27.28/91.06$ & $26.10/92.52$ & $25.60/91.38$ \\ 
& TIP-IM & $19.72/88.60$ & $27.96/86.32$  & $22.78/89.65$ & $28.64/90.45$  & $24.66/91.38$ & $26.80/91.52$ & $26.74/90.56$ & $25.60/91.37$ \\ 
\bottomrule
\end{tabular}
}
\vspace{-0.05in}
\label{table: main results full}
\end{table*}

\section{Other Experimental Details}
\label{appendix: detailed experimental settings}

In addition to the main experimental setup described in Section \ref{section: experiments}, we lay out all other necessary implementation details for completeness.

\shortsection{Configuration} 
For all the other non-Trackee identities, $10\%$ of their facial images are used as query images, while the remaining images are added to the gallery database of the FR model employed by Tracker. 
During the process of the AFR-protected image generation, we use the Adam optimizer for the gradient descent steps in Equation \ref{equation: gradient descent}, where the hyperparameters $\beta_1$ and $\beta_2$ are set to $0.9$ and $0.999$, respectively. For the face verification experiments, in addition to the trackee's other identities, each identity selects one image as the query image.

\shortsection{Dataset} 
The FaceScrub dataset contains $106,863$ photos of $530$ celebrities with annotated names and genders, which are retrieved from the Internet and captured in real-world scenarios.
The PubFig dataset contains $58,797$ images of $200$ individuals with annotated ages, races, and occupations, which are largely various in pose, lighting, and scene. The UMDFaces dataset contains $367,888$ images of $8,277$ individuals with annotated genders and postures, where faces are located by human-curated bounding boxes. We use the CelebA-HQ dataset organized by identity~\cite{na2022unrestricted}, which contains a total of $307$ individuals and $5,478$ images, with each identity having more than 15 images.

\shortsection{Feature Extractor} 
For the FR model employed by Tracker, we consider $4$ pre-trained facial feature extractors in our evaluation, including MobileFace~\cite{chen2018mobilefacenets} trained on MS1MV2 using MobileFaceNet, WebFace~\cite{yi2014learning} that is trained on CASIA-WebFace using Inception ResNet~\cite{yi2014learning}, 
VGGFace~\cite{cao2018vggface2}, which is pre-trained on VGG-Face2 using Inception ResNet~\cite{yi2014learning}, and MagFace~\cite{meng2021magface} trained on MS1MV2 using MagFace loss and ResNet. 
Note that these feature extractors are set to be different from those used for generating AFR protections. 

\section{Additional Experiments}

In this section, we conduct additional experiments to study DynTracker's strength in breaking existing AFR protection schemes and the effectiveness of our DivTrackee in enhancing protection efficacy against dynamic FR strategies while preserving image visual quality.

\subsection{Evaluation of Adversarial Noise-Based AFR}
\label{appendix: full table results}

In Section \ref{subsection: main results}, we evaluate the efficacy of different generative-based AFR protection schemes against static and dynamic FR strategies. To more comprehensively examine the effectiveness of our DynTracker, we further evaluate the performance of existing adversarial noise-based AFR methods, including PGD~\cite{madry2018towards}, MI-FGSM~\cite{dong2018boosting}, TI-DIM~\cite{dong2019evading}, and TIP-IM~\cite{yang2021towards}, under the same experimental settings. 
The evaluation results are demonstrated in Table \ref{table: main results full}, which again confirms the limited protection success of existing AFR protection schemes against our DynTracker, regardless of Tracker initially holding a clean or protected Trackee image.

\subsection{Influence of Target Text prompt}
\label{appendix: influence of text prompt}

\begin{figure}[!t]
    \centering
    \begin{subfigure}[b]{0.37\textwidth}
        \includegraphics[width=\linewidth]{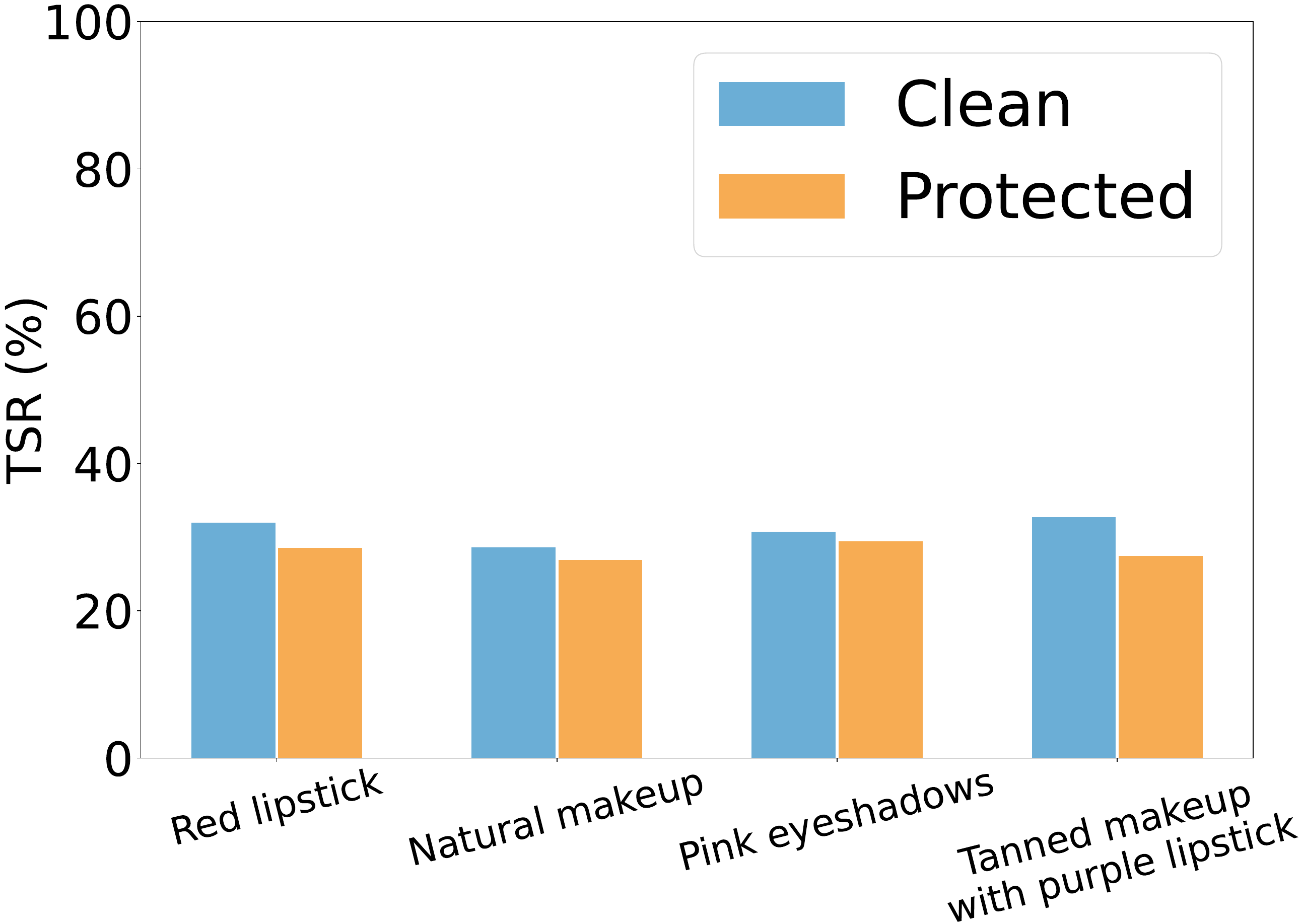}
        \caption{FaceScrub}
        \label{figure: text prompt FaceScrub}
    \end{subfigure}
    \hspace{0.2in}
    \begin{subfigure}[b]{0.37\textwidth}
        \includegraphics[width=\linewidth]{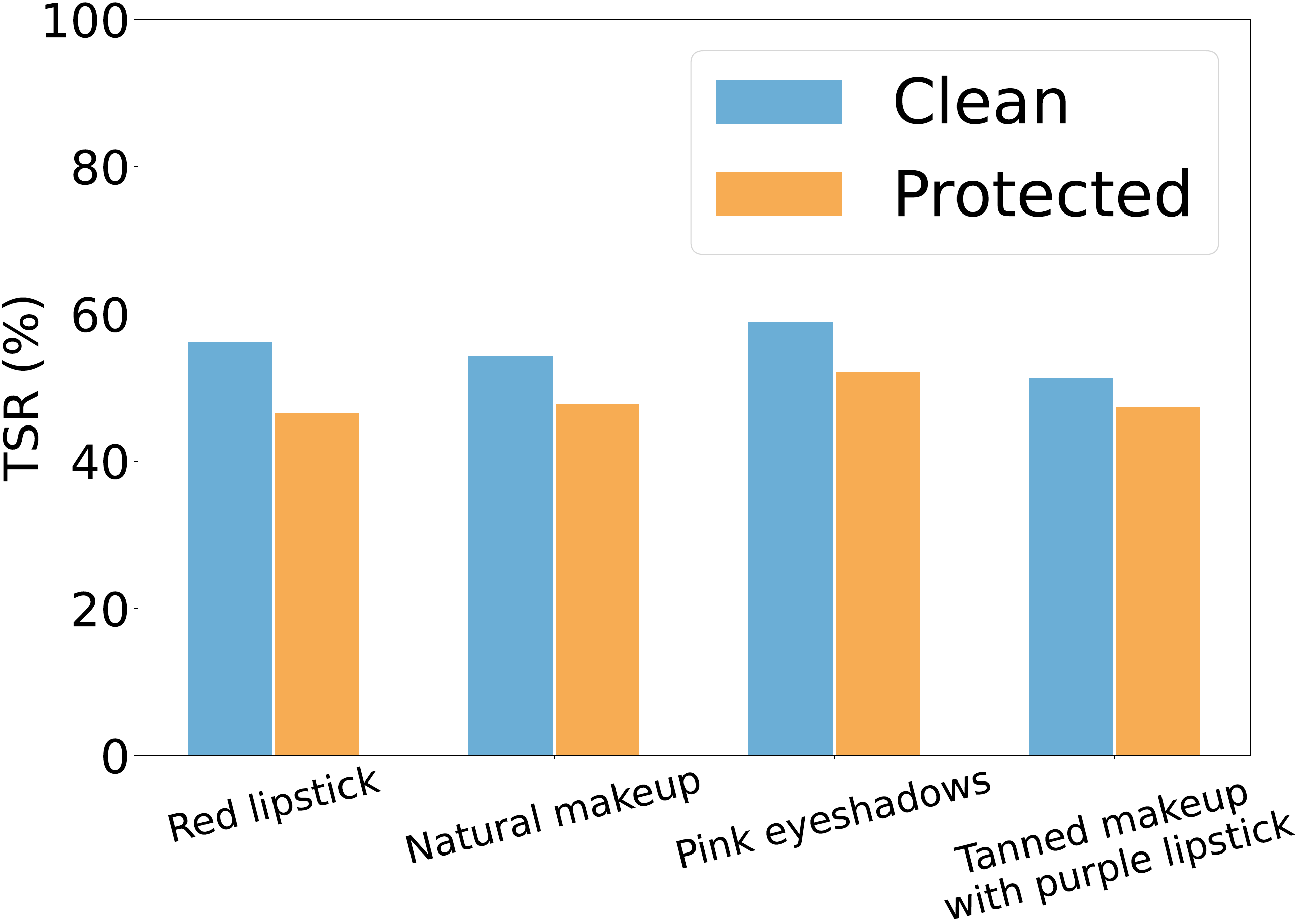}
        \caption{CelebA-HQ}
        \label{figure: text prompt CelebA-HQ}
    \end{subfigure}
    \vspace{-0.1in}
    \caption{Comparisons of tracking success rates of DynTracker across various target text prompts, which are used to guide the generation process of AFR-protected Trackee images, on two face datasets: (a) FaceScrub, and (b) CelebA-HQ. 
    }
    \vspace{-0.05in}
    \label{figure: text prompt}
\end{figure}

\begin{figure}[!t]
    \centering
    \includegraphics[width=0.7\textwidth, height=5.5cm]{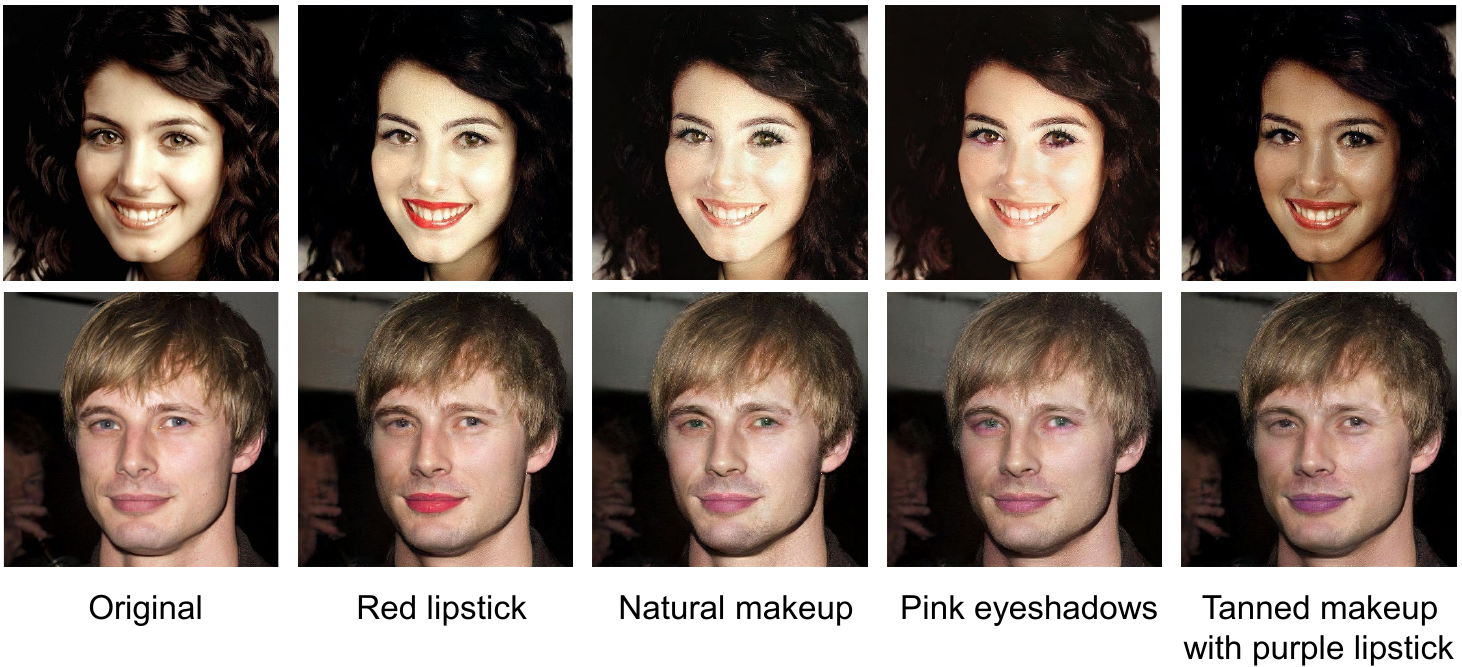}
    \vspace{-0.1in}
    \caption{Visualizations of the generated AFR-protected Trackee facial images on CelebA-HQ across different text prompts.
    }
    \vspace{-0.05in}
    \label{figure: text prompt example}
\end{figure}

We investigate the impact of different target text prompts $p_{\mathrm{targ}}$ on the efficacy of DivTrackee. 
Figure \ref{figure: text prompt} indicates that for both the ``Clean'' and ``Protected'' cases, the tracking success rates against our DynTracker exhibit slight variations across different text prompts, while the visualizations depicted in Figure \ref{figure: text prompt example} illustrate that text prompts primarily influence the final makeup effect of the protected images generated by DivTrackee, largely aligned with the expectations of Trackee for AFR in terms of visual quality.

\subsection{Quantitative Results on Visual Quality}
\label{appendix: quantitative comparison results}

We adopt the metrics of \emph{Structural Similarity Index Measure} (SSIM)~\cite{WBSS04} and \emph{Peak Signal to Noise Ratio} (PSNR) to measure the visual quality of generated protected facial images quantitatively.
Specifically, SSIM measures the similarity between two images by comparing the conditions of luminance, contrast, and structure; meanwhile, PSNR defines the similarity as the peak signal-to-noise ratio between two images
Besides, a higher value of SSIM/PSNR indicates better visual quality. In Table \ref{table: visual quality}, while our method yields slightly lower SSIM and PSNR values compared to other approaches, the visual quality remains acceptable for users. We present the protected images generated using various prompts in Figure \ref{figure: text prompt example}. We can observe that the generated images preserve the original facial features while incorporating the appropriate makeup effects.


\begin{table}[!t]
\centering
\caption{Quantitative comparisons of visual quality between different AFR methods on FaceScrub.}
\vspace{-0.1in}
    \begin{tabular}{l | c c c c}
        \toprule
        \textbf{Method} & \textbf{AMT-GAN} & \textbf{Clip2Protect} & \textbf{DiffAM} & \textbf{DivTrackee} \\
        \midrule
        SSIM $(\uparrow)$ & 0.43 & 0.38 & 0.40 & 0.38 \\
        PSNR $(\uparrow)$ & 12.09 & 11.59 & 11.78 & 11.55 \\
        \bottomrule
    \end{tabular}
\label{table: visual quality}
\end{table}

\end{document}